\documentclass[conference]{IEEEtran}
\IEEEoverridecommandlockouts
\usepackage{amsmath,amssymb,amsfonts}
\usepackage{paralist}
\usepackage{algorithm}
\usepackage{algpseudocode}
\usepackage{graphicx}
\usepackage{textcomp}
\usepackage{xcolor}
\usepackage{tikz}
\def\BibTeX{{\rm B\kern-.05em{\sc i\kern-.025em b}\kern-.08em
    T\kern-.1667em\lower.7ex\hbox{E}\kern-.125emX}}

\usepackage{amsmath,amsfonts,bm}
\usepackage{amsthm}
\usepackage{mathtools}

        {\hspace*{\fill}$\Box$\par}

\newtheorem{innercustomgeneric}{\customgenericname}
\providecommand{\customgenericname}{}
\newcommand{\newcustomtheorem}[2]{%
  \newenvironment{#1}[1]
  {%
  \renewcommand\customgenericname{#2}%
  \renewcommand\theinnercustomgeneric{##1}%
  \innercustomgeneric
  }
  {\endinnercustomgeneric}
}

\newcustomtheorem{customthm}{Theorem}
\newcustomtheorem{customlemma}{Lemma}
\newcustomtheorem{customproblem}{Problem}

\DeclarePairedDelimiterX{\infdivx}[2]{(}{)}{%
  #1\;\delimsize\|\;#2%
}

\def\eqref#1{equation~\ref{#1}}

\def\1{\bm{1}}

\def\rva{{\mathbf{a}}}
\def\rvb{{\mathbf{b}}}
\def\rvc{{\mathbf{c}}}

\def\rvs{{\mathbf{s}}}

\def\rvw{{\mathbf{w}}}
\def\rvx{{\mathbf{x}}}
\def\rvy{{\mathbf{y}}}

\DeclareMathAlphabet{\mathsfit}{\encodingdefault}{\sfdefault}{m}{sl}
\SetMathAlphabet{\mathsfit}{bold}{\encodingdefault}{\sfdefault}{bx}{n}

\def\gA{{\mathcal{A}}}

\def\gC{{\mathcal{C}}}
\def\gD{{\mathcal{D}}}
\def\gE{{\mathcal{E}}}

\def\gG{{\mathcal{G}}}
\def\gH{{\mathcal{H}}}

\def\gL{{\mathcal{L}}}

\def\gN{{\mathcal{N}}}
\def\gO{{\mathcal{O}}}
\def\gP{{\mathcal{P}}}

\def\gR{{\mathcal{R}}}
\def\gS{{\mathcal{S}}}

\def\gV{{\mathcal{V}}}

\def\gX{{\mathcal{X}}}

\newcommand{\R}{\mathbb{R}}

\DeclareMathOperator*{\argmin}{arg\,min}

\usepackage{float}
\usepackage{diagbox}
\usepackage{adjustbox}
\usepackage{slashbox}
\usepackage{subcaption}
\usepackage{xspace}
\usepackage{url}
\usepackage[inline]{enumitem}
\usepackage[colorlinks,allcolors=red]{hyperref}

\newcommand{\delaczero}{\text{D.MCA-0}\xspace}
\newcommand{\delac}{\text{D.MCA}\xspace}
\newcommand{\gensqout}{\text{Gen$^2$Out}\xspace}

\usepackage{multibib}
\newcites{app}{Appendix References}%

\newcommand{\hide}[1]{}

\makeatletter
\newcommand\footnoteref[1]{\protected@xdef\@thefnmark{\ref{#1}}\@footnotemark}
\makeatother

\renewcommand{\algorithmicrequire}{\textbf{Input:}}
\renewcommand{\algorithmicensure}{\textbf{Output:}}

\newcommand{\cbit}{\begin{compactitem}}
\newcommand{\ceit}{\end{compactitem}}
\newcommand{\cben}{\begin{compactenum}}
\newcommand{\ceen}{\end{compactenum}}

\newcommand{\beq}{\begin{equation}}
	\newcommand{\eeq}{\end{equation}}

\newcommand{\bit}{\begin{itemize}}
	\newcommand{\eit}{\end{itemize}}
\newcommand{\ben}{\begin{enumerate}}
	\newcommand{\een}{\end{enumerate}}

\newcounter{x}

\algrenewcommand\algorithmicrequire{\textbf{Input:}}
\algrenewcommand\algorithmicensure{\textbf{Output:}}

\begin{document}

\title{\LARGE{\delac: Outlier \underline{D}etection with Explicit \underline{M}icro-\underline{C}luster \underline{A}ssignments}}
\author{\IEEEauthorblockN{Shuli Jiang}
\IEEEauthorblockA{\textit{Carnegie Mellon University} \\
shulij@andrew.cmu.edu}
\and
\IEEEauthorblockN{Robson L. F. Cordeiro}
\IEEEauthorblockA{\textit{University of São Paulo} \\
robson@icmc.usp.br}
\and
\IEEEauthorblockN{Leman Akoglu}
\IEEEauthorblockA{\textit{Carnegie Mellon University} \\
lakoglu@andrew.cmu.edu}
}

\maketitle

\begin{abstract}
    How can we detect outliers, both scattered and clustered, and also explicitly assign them to respective micro-clusters, without knowing apriori how many micro-clusters exist?
    How can we perform both tasks in-house, i.e., without any post-hoc processing, so that both detection and assignment can benefit simultaneously from each other?
    Presenting outliers in separate micro-clusters is informative to analysts in many real-world applications.
    However, a na\"ive solution based on post-hoc clustering of the outliers detected by any existing method suffers from two main drawbacks:
    (a) appropriate hyperparameter values are commonly unknown for clustering, and most algorithms struggle with clusters of varying shapes and densities;
    (b) detection and assignment cannot benefit from one another.
    In this paper, we propose \delac to \underline{D}etect outliers with explicit \underline{M}icro-\underline{C}luster \underline{A}ssignment. 
    Our method performs both detection and assignment iteratively, and in-house, by using a novel strategy that prunes entire micro-clusters out of the training set to improve the performance of the detection.
    It also benefits from a novel strategy that avoids clustered outliers to mask each other, which is a well-known problem in the literature.
    Also, \delac is designed to be robust to a critical hyperparameter 
    by employing a hyperensemble ``warm up'' phase.
    Experiments performed on $16$ real-world and synthetic datasets demonstrate that \delac outperforms $8$ state-of-the-art competitors, especially on the explicit outlier micro-cluster assignment task.
\end{abstract}


\section{Introduction}
\label{sec:intro}

Outlier detection aims to identify rare or unusual instances in the data that deviate from the majority.
In many practical domains outliers can form groups or (micro-)clusters, which can be seen as anomalous patterns. Examples include bots or malware under the same command-control in network intrusion, coordinated attacks from multiple user accounts toward spreading fake news in social platforms, opportunistic fraud/scam/etc. that spread by word-of-mouth, adversarial schemes that exploit the same loophole or vulnerability in a system (e.g., medical insurance), to name a few.

Autonomous outlier detection systems are rarely used in the real world,
where it is typical for the flagged outliers to go through a verification/vetting process by a domain analyst, especially when the cost of false positives is high, such as shutting down credit cards, blacklisting certain software or user accounts from access, charging a physician or hospital by fraud, etc. It is exactly in such scenarios that reporting  micro-clusters of outliers, if any, is useful for the domain analyst whose job may be to go through hundreds of such alerts a day. Rather than inspecting each outlier one by one, a succinct presentation of the outliers by groups could speed up sense-making, characterization and ultimately decision making, where e.g. the analyst can take the same troubleshooting action for all outliers in the same cluster, or quickly ignore all in the same cluster provided a few have already been found to be false positives or semantically uninteresting outliers.

\begin{figure}[!tbp]
  \centering
  \subfloat[\delac]{\includegraphics[width=0.33\linewidth]{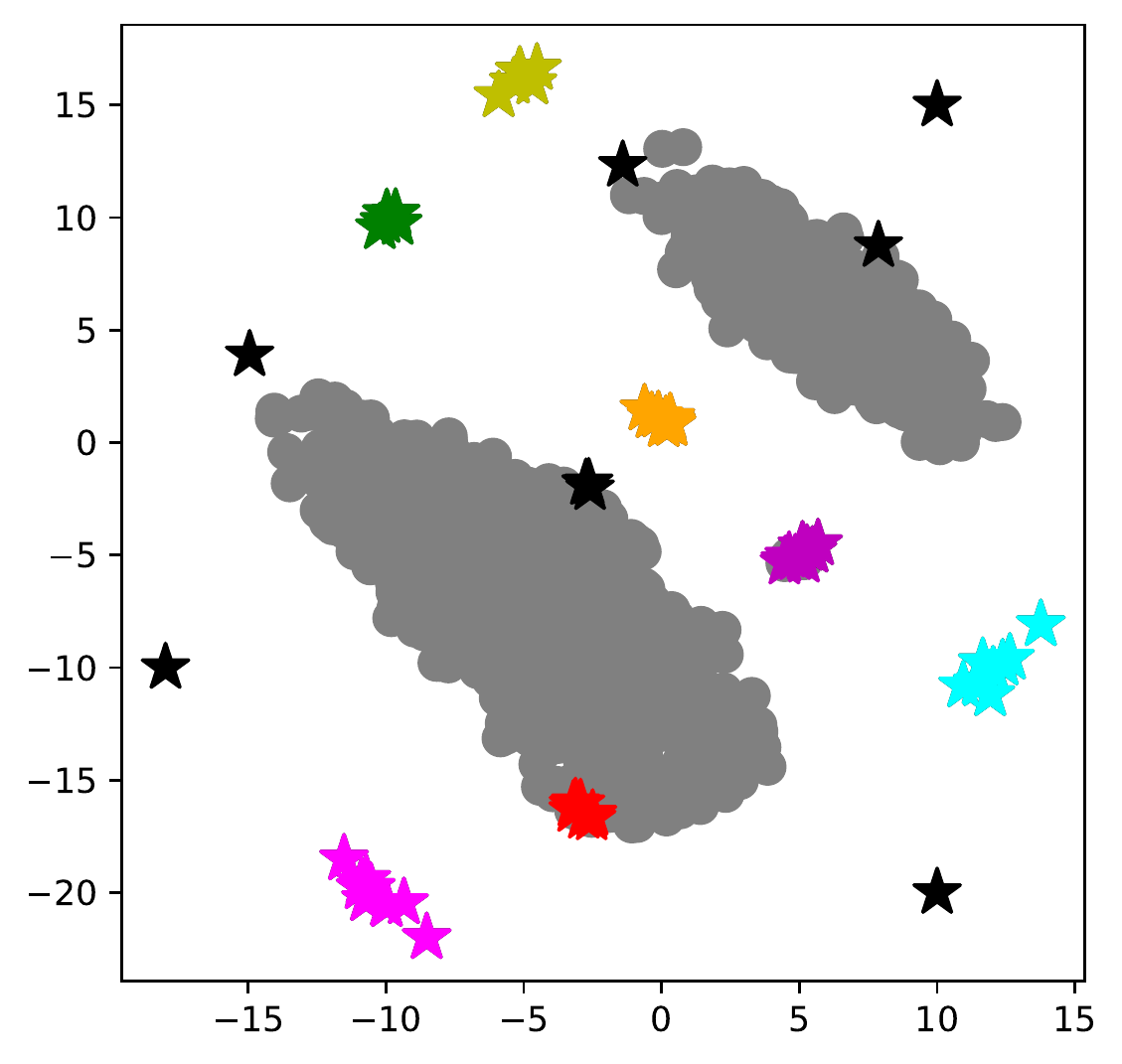}}
\subfloat[iForest+X-means]{\includegraphics[width=0.33\linewidth]{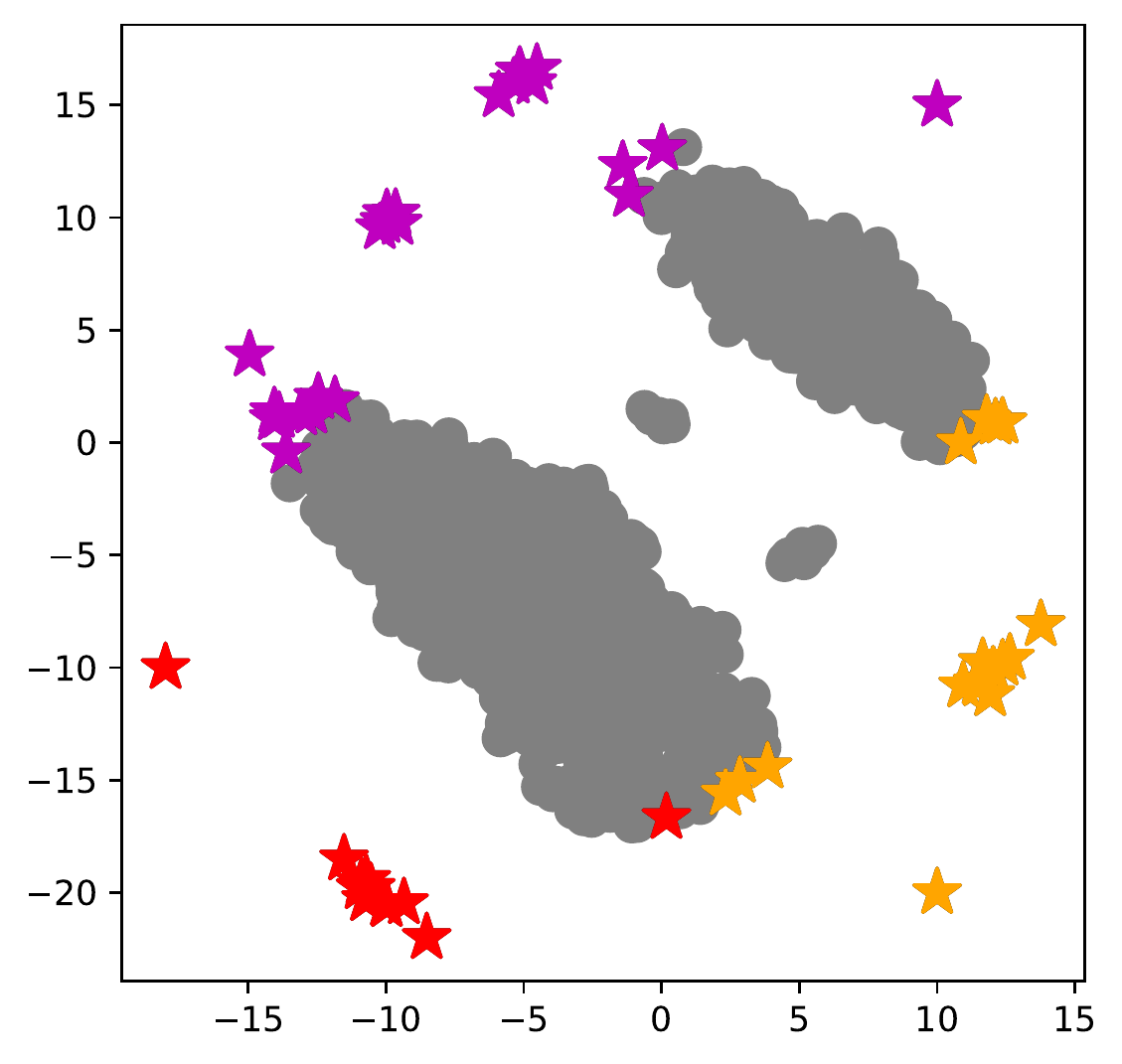}}
  \subfloat[\gensqout]{\includegraphics[width=0.33\linewidth]{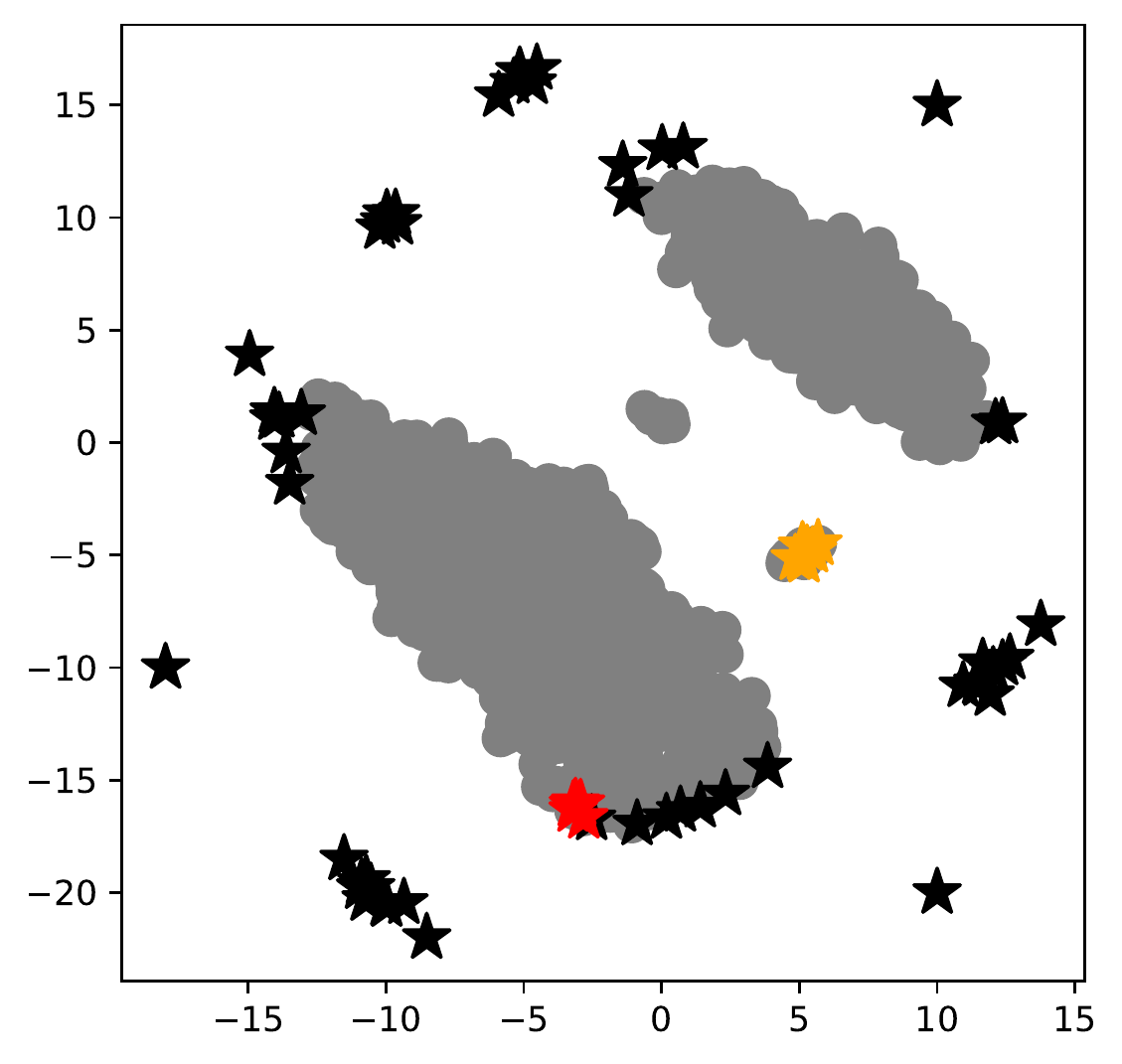}}
  
  \caption{Example dataset with two inlier modalities (gray dots), with scattered and clustered outliers. (a) Proposed \delac effectively flags the outliers (stars) and also explicitly assigns to micro-clusters (colored differently). (b) Two-stage solutions, detection+post hoc clustering, are not as effective. (c) SOTA baseline underperforms, often missing interior outliers.}
  \vspace{-0.05in}
  \label{fig:eg}
\end{figure}

The specific problem we consider in this work is then to not only identify the outlier instances (both clustered and scattered) but at the same time, output existing outlier micro-clusters explicitly. We refer to the latter task as outlier \textit{micro-cluster assignment} (e.g. see Fig. \ref{fig:eg}(a)).
Our goal is to design an algorithm that addresses both problems \textit{in-house} as a result of its inherent detection mechanism, rather than {\em post-hoc} where one employs a clustering algorithm upon first detecting the outliers.
Such straightforward  approaches are arguably far from trivial to get right due to multi-pronged challenges. First, the detection algorithm should be very effective as those outliers are fed downstream to clustering. Second, clustering itself is a non-trivial task with small, i.e. micro, clusters of possibly varying size and density, where further the number of outlier clusters are not known apriori, all of which relate to the challenge of hyperparameter selection. Finally, a two-stage approach puts a barrier between the detection and assignment tasks that prevents them from potentially benefiting  from each other. We compare to various two-stage baselines in experiments, using SOTA detectors such as Isolation Forest \cite{liu2008isolation_forest} and LOF \cite{breunig2000LOF}, paired with parameter-free clustering algorithms such as X-means \cite{pelleg2000x} and OPTICS \cite{ankerst1999optics} (e.g. see Fig. \ref{fig:eg}(b)).

The vast body of prior work is for detecting outliers only, without special focus on clustered outliers. 
 In principle several of those detectors are capable of spotting outliers that are present in micro-clusters, provided suitable hyperparameter values, however they do not specify which outliers are scattered versus clustered. There exist work clustering detected outliers which is post-hoc and mainly focuses on explanation \cite{journals/datamine/MachaA18}. 
 Some work have addressed  detecting clustered or collective outliers specifically \cite{liu2010SCiForest, silva2021callout,chandola2009anomaly}, however without explicitly assigning them to their respective clusters.
The only work (to our best knowledge), \gensqout~\cite{lee2021gen2out}, that considers explicit outlier assignment is based on the aforementioned na\"ive two-stage pipeline: it detects outliers first and then applies the post-processing clustering algorithm DBSCAN~\cite{ester1996dbscan} on the detected outliers. It relies heavily on the accuracy of the detector which falls short to spot outliers in the interior of the data manifold (e.g. see Fig. \ref{fig:eg}(c)). See Appendix~\ref{sec:related_works} for detailed related work.



In this paper, we simultaneously address the outlier \underline{D}etection and explicit outlier \underline{M}icro-\underline{C}luster \underline{A}ssignment tasks. Our proposed \delac alternatingly refines the micro-clusters and the flagged outliers, aiming to synergize them to mutually benefit from one another.
In a nutshell, \delac builds on the SOTA iNNE detector \cite{bandaragoda2018iNNE} which subsamples a set of $\psi$ points, creates a hyperball centered at each one, and scores outlierness as relative distance to those representatives. The working assumption is that 
the subsample contains inliers with high probability, and hence the hyperballs are representative of the normal data distribution. Note that iNNE originally cannot produce explicit micro-cluster assignments.

Our proposed \delac improves this base algorithm in two key aspects. 
Firstly, it employs an iterative pruning strategy where the top scoring points at each iteration are pruned. In effect, this improves the probability that the subsampled points in the next round are more likely to be inliers, which lead to better outlier detection.  
Second, \delac employs a hyperensemble ``warm up'' phase that makes it more robust to the choice of its hyperparameter, namely the subsample size $\psi$, which at the same time also helps alleviate the issue of ``masking''; when two or more points from an outlier cluster are sampled, they mask/hide one another and cause all points in the micro-cluster to be deemed inliers (i.e. false negatives). 
We summarize our main contributions as follows.
\cbit
\item {\bf Outlier detection and micro-cluster assignment:} We consider the two-pronged problem of (scattered and clustered) outlier detection and explicit outlier micro-cluster assignment, and propose a new algorithm called \delac to address both problems simultaneously.

\item {\bf Tackling various challenges:} \delac exhibits novel procedures toward $(i)$ effective outlier pruning -- which specifically aims to avoid pruning false positives (i.e. inliers),
$(ii)$ reduced ``masking'' effect -- which carefully regulates the subsample size\footnote{Note that a large sample is more likely to induce ``masking'', leading to false negatives (i.e. missed outlier micro-clusters),  whereas a small sample would contain insufficient representatives, leading to false positives.} and prunes true positive micro-clusters, and    
$(iii)$ improved robustness to hyperparameter settings -- which employs a hyperensemble strategy as a ``warm up'' phase.

\item {\bf A synergistic solution:} \delac's detection and micro-cluster assignment steps work together in synergy under a unified algorithm, rather than two independent components: the pruning strategy reveals micro-clusters as a by-product, and true-positive micro-clusters are pruned to improve the subsample and hence the detection quality.

\item {\bf Effectiveness:} Through extensive experiments on synthetic datasets, real-world datasets injected with known clustered outliers, as well as benchmark real-world datasets, we show that \delac sets the state-of-the-art, significantly outperforming (to our knowledge) the only SOTA baseline \gensqout as well as a number of two-stage baselines with paired detector and post-hoc clustering, especially in the micro-cluster assignment task.
\ceit
{\bf Reproducibility.~} All our source code and the datasets used in the experiments are publicly shared at \url{https://github.com/11hifish/D.MCA}.

\section{Preliminaries}
\label{sec:prelim}

{\bf Notation}
 See Appendix~\ref{subsec:notations}.

{\bf Problem} Given a point-cloud dataset $\gD$, our goal is to perform the following two tasks: 

1) {\em Detection:} compute $\rvx_i \in \gD, \forall i \in [n]$ an outlier score $s(\rvx_i)$; the larger, the more anomalous and 

2) {\em Assignment:} explicitly output outlier micro-clusters $\gC_1,\dots, \gC_m$, where $m$ is apriori unknown to the algorithm.

{\bf iNNE Outlier Detector} 
\delac uses individual models of Isolation using Nearest Neighbor Ensemble (iNNE)~\cite{bandaragoda2018iNNE} as the base outlier detector.
See Appendix~\ref{subsec:inne_algo} for more details.

\section{Proposed Approach}
\label{sec:method}
The section is divided into two parts. We first present the \delaczero algorithm (Algorithm~\ref{alg:delac_zero}), a sequential ensemble that alternates between detection and assignment at each iteration. However, \delaczero is sensitive to an important hyperparameter $\psi$, the subsample size, inherited from the iNNE base detector. Next we present the proposed \delac (Algorithm~\ref{alg:delac}), a two-phase algorithm that builds a \textit{hyperensemble} based on varying values of $\psi$ in the first phase prior to employing \delaczero in the second phase, which is more robust to $\psi$ compared to \delaczero.

\subsection{Proposed \delaczero}
\label{subsec:delac_zero}

The motivation of performing detection and assignment interactively across iterations in a sequential ensemble is two-fold: ($i$) to obtain a better assignment of the outliers based on improved detection, and ($ii$) to leverage the identified micro-clusters to improve the detection performance. 

At a high level, \delaczero builds upon iNNE to detect outliers, which outputs an outlier score $s$ for each instance in $\gD$ and a set of selected centers $\gR_{\psi}$ as inlier representatives per iteration. \delaczero keeps a running average of the outlier scores across iterations. We choose iNNE as the base model as it is fast and efficient, able to detect non-axis aligned outliers, adapts to data with varying density,  and produces explicit subsamples $\gR_{\psi}$ that our algorithm makes use of. \delaczero keeps track of the assignment of outlier micro-clusters by a weighted neighbor graph $\gG = (\gD, \gE)$ where $\gE = (\rvx, \rvy), \forall \rvx, \rvy \in \gD$. The weight of an edge $(\rvx, \rvy) \in \gE$ denotes the number of times both instances $\rvx, \rvy \in \gD$ receive high outlier scores and are marked as ``neighbors''. \delaczero updates the weight of $\gG$ at each iteration, and outputs the final outlier micro-clusters $\gC = \{\gC_1, \gC_2, \dots\}$ based on the connected components of  $\gG$  (Procedure \textsf{\small{FindClusters}} in Appendix~\ref{sec:find_clusters}).

In what follows, we present four major steps of \delaczero during a single iteration in the sequential ensemble: \textit{1)} finding micro-cluster representatives via sampling, \textit{2)} filtering out false positives from micro-cluster representatives toward pruning, \textit{3)} updating the neighbor graph $\gG$ and \textit{4)} pruning true micro-cluster representatives and their neighbors. We remark that steps \textit{1)} and \textit{3)} are geared toward $(i)$ to obtain outlier micro-clusters and steps \textit{2)} and \textit{4)} are geared toward $(ii)$ to improve detection performance. Notice how these steps are interleaved, creating  synergy between the two tasks. 

\subsubsection{Finding Micro-cluster Representatives via Maximin Sampling}

To find outlier micro-clusters, the algorithm first applies Maximin Sampling~\cite{ibrahim2020maximin_sampling} on outliers that receive $p$ topmost outlier scores, i.e. points in $\gH$-top (Line 7), to get an outlier representative set $\gH$ which contains at least one representative per micro-cluster (Line 8). 
At each iteration, Maximin Sampling samples an unselected point in the dataset that has the maximum projection to the set of already sampled points.\footnote{Projection is the minimum distance from a point to a set.} When there are well-separated outlier micro-clusters, Maximin Sampling guarantees that we sample at least one point per micro-cluster. Furthermore, if all micro-clusters have one subsample already by Maximin Sampling, the projection of the next selected points is expected to have a large decrease, which gives us an estimation of the number of outlier micro-clusters. We stop sampling after such a large decrease so that $\gH$ does not include an unnecessarily large number of points from the micro-clusters, to improve algorithm efficiency. We later use such outlier representatives as ``anchor points'' to identify different micro-clusters.

\begin{algorithm}[]
\caption{\textsc{\delaczero}}
\label{alg:delac_zero}
\begin{algorithmic}[1]
\Require data $\gD \in \R^{n \times d}$, subsample size $\psi$, num. iterations $t$ (default: 100),  num. check-points $p$ (default: 0.1$n$)
\Ensure outlier scores $\bar{\rvs} \in [0, 1]^n$, explicit set of outlier micro-clusters $\gC=\{\gC_1, \gC_2, \ldots\}$, neighbor graph $\gG$
\State Outlier scores $\bar{\rvs} \leftarrow {\bf 0} \in \R^{n}$ (init.)
\State Neighbor graph $\gG \leftarrow (\gV = \gD, \gE = \emptyset)$ (init.)
\State Clean set $\gR \leftarrow \gD$ (init.)
\For{$i = 1, 2, \dots, t$}
\State $\rvs^{(i)}, \text{centers } \gR_{\psi}^{(i)} \leftarrow $ i\textsc{NNE} (train: $\gR$, test: $\gD$, $\psi$: $\psi$) \label{delac:line:inne}
\State $\bar{\rvs} \leftarrow
(\bar{\rvs} * (i-1) + \rvs^{(i)}) / i$
\State High-score set $\gH\text{-top} \leftarrow \{\rvx \in \gD: \bar{\rvs}[\rvx] \text{ among top $p$}\}$ 

\State Representative set $\gH \leftarrow $ \textsf{\small{MaximinSampling}}($\gH\text{-top}$) \label{delac:line:mm_sampling}
\State maximum radius $r_{\max}\leftarrow \max_{\rvx \in \gH} \min_{\rvc \in \gR_{\psi}^{(i)}} \|\rvx - \rvc\|_2$
\State $\gA \leftarrow \emptyset$ $\;\;\blacktriangleright$ \texttt{\# areas under} ``\texttt{clothes-lines}''
\State Sorted distances $\gL \leftarrow \emptyset$ 
\For{$\rvx \in \gH \cup R_{\psi}^{(i)}$} \Comment{\texttt{\#clothes-lines}}
\label{delac:line:clothes_lines}
 
    \State distance $\text{Dist}[\rvx,\rvy] \leftarrow \|\rvy - \rvx\|_2, \forall \rvy \in \gD$ \label{delac:line:all_distance}
    \State $\gL[\rvx] \leftarrow$ sorted Dist.s to $\rvx$ from smallest to largest 
    \State $\rva \leftarrow$ sorted average neighboring outlier scores \label{delac:line:sorted_avg_score}
    \State $\gA[\rvx] \leftarrow $ 
    weighted sum of $\rva$  based on $\gL[\rvx]$ and  $r_{\max}$ (See  Eq.~(\ref{eq:weighted_sum}))
\EndFor \label{delac:line:clothes_lines_end}
   \State \texttt{\# Decide whether $\rvx$ is a true outlier}
\State Candidate true set $\gH_\text{c} \leftarrow \{\rvx \in \gH: \gA[\rvx] > \text{mean}(\gA)\}$
\State Pruning set $\gP \leftarrow \emptyset$
\For{$\rvx \in \gH_\text{c}$}
    \State threshold $\tau_n \leftarrow$ \textsf{\small{FindFirstPeak}}($\gL[\rvx]$)
    \State \texttt{\# Find neighbors of $\rvx$ to prune}
    \State Neighbors $\gN \leftarrow \{\rvy \in \gD: \|\rvx - \rvy\|_2 < \tau_n\}$
    \State $\gG \leftarrow$ increase edge weight ($\rvx, \rvy$) by 1, $ \forall \rvy \in \gN$
    \State $\gP \leftarrow \gP \cup \gN$
\EndFor
\State Clean set $\gR \leftarrow \gD \setminus \gP$
\EndFor
\State $\gC \leftarrow$ \textsf{\small{FindClusters}}($\gG$) \Comment{\texttt{\# Procedure~\ref{alg:find_clusters}}} \\
\Return $\bar{\rvs}, \gC, \gG$
\end{algorithmic}
\end{algorithm}
\setlength{\textfloatsep}{0.15in}

\subsubsection{Filtering out False Positives from Micro-cluster Representatives toward Pruning}

\delaczero prunes outlier micro-clusters found at each iteration to improve the detection performance. Pruning can reduce one major drawback of iNNE at the presence of outlier micro-clusters, called ``masking''. Masking occurs when two or more points from the same outlier micro-cluster is subsampled into $\gR_{\psi}$, which causes the members of the micro-cluster to be masked or hidden as inliers. Hence, training with a cleaner set with fewer outlier micro-clusters decreases the probability of ``masking'' and leads to better detection performance.

However, na\"ively pruning points with the topmost outlier scores is not effective as this leads to also pruning false positives, i.e. inlier points that receive high outlier scores, which leads to even worse performance than iNNE (e.g. Fig.~\ref{fig:motivation_puning} in the Appendix). Therefore, we perform pruning by first deciding whether an outlier representative in $\gH$ is a false positive, following a
careful procedure that leverages what we call
``clothes lines'' (Lines~\ref{delac:line:clothes_lines}-\ref{delac:line:clothes_lines_end}), and prune only based on points in $\gH$ that are estimated to be true outliers. 

The idea is that false positive points are closer to many inlier points compared to those true outliers by definition, and thus the average score of the neighbors of false positive points is expected to be lower as compared to that of the true outliers. 
We describe the details of clothes-lines, i.e. distinguishing false positives from true positives next.

\begin{figure}[]
    \centering
    \includegraphics[width=0.8\linewidth]{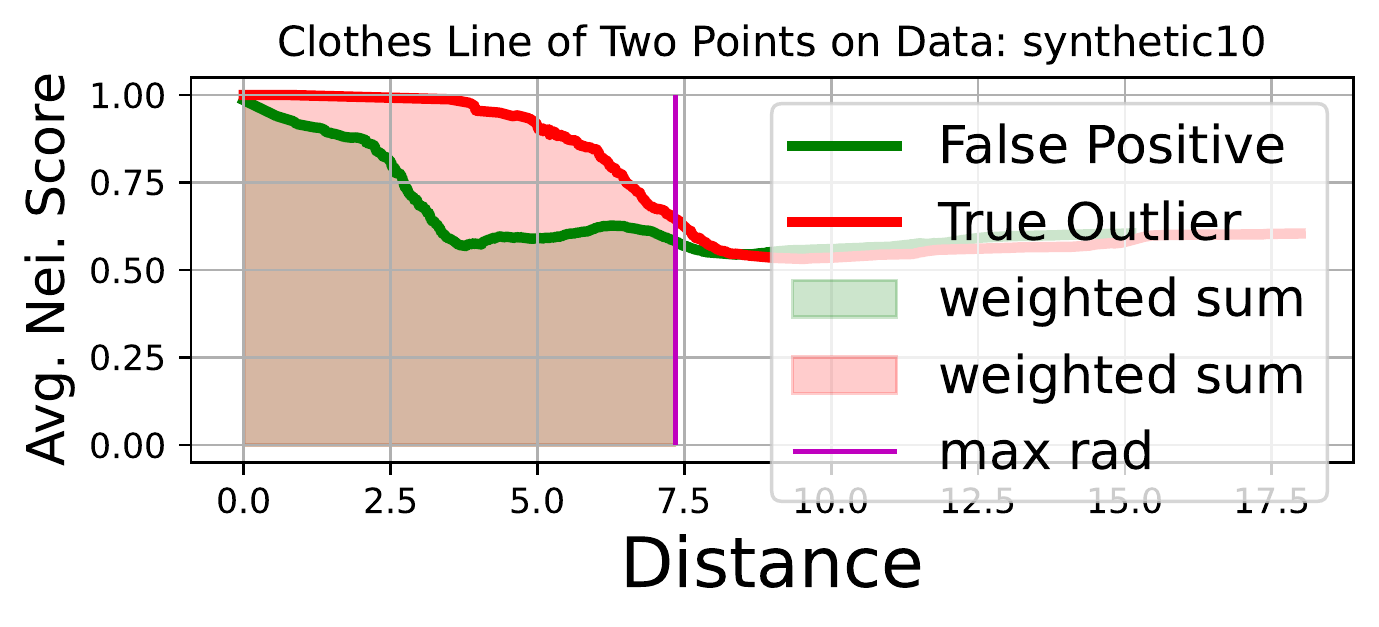}
    \vspace{-0.05in}
    \caption{Example clothes-line of an outlier (red curve) and an inlier (green curve). For an outlier (i.e. true positive), average neighboring scores (y-axis) by distance (x-axis) remains high and flat, and hence looks like a ``clothes-line'' while the curve for a false positive inlier drops relatively quickly, even if they both receive a high score initially (as shown at distance 0).}
    \label{fig:clothes_line_example}
\end{figure}

{\bf Computing Clothes-lines and Weighted Sums.~}
To construct the clothes-line for a point $\rvx$, we first compute the distances from all points in $\gD$ to $\rvx$ (Line 13), and sort the distances from the smallest to the largest to get $\gL[\rvx] \in \R^{n}$ (Line 14). We then compute the average outlier score of the neighbor points of $\rvx$ with an increasing distance to $\rvx$, i.e. we compute $\rva \in \R^{n}$ (Line 15), such that $\rva_i = \sum_{\rvy: \|\rvy - \rvx\|\leq \gL[\rvx]_i} \bar{\rvs}[\rvy] / i$, $\forall i \in [n]$.
The clothes-line of $\rvx$ is then a curve on a 2D plot where the x-axis is $\gL[\rvx]$  and the y-axis is $\rva$. We illustrate in Fig.~\ref{fig:clothes_line_example} example clothes-lines at iteration 10 for an outlier (in red) and an inlier (in green) in \texttt{synthetic10} (one of our datasets with ground-truth, see Appendix~\ref{ssec:datasets_appendix}).

Intuitively, for an outlier, the average outlier score of points by increasing distance is large and remains large\footnote{This is the case for both scattered and clustered outliers, where for the former the average includes only the point itself for a large radius around it.}, as such the curve looks flat like a clothes-line. In contrast, for an inlier (i.e. false positive), the average tends to drop fairly quickly (i.e. within a shorter distance away).
Then, we use the area under the clothes-lines (shaded areas in Fig.~\ref{fig:clothes_line_example}) to distinguish between true and false positives, which is 
computed through a weighted sum of the outlier scores, 
where the weights are the distances to neighboring points (Line 16). Specifically, the weighted sum for $\rvx$ is given as 
\begin{align}
\label{eq:weighted_sum}
    \gA[\rvx] := \sum_{i: \gL[\rvx]_i \leq r_{\max}} \rvw_i \cdot (\gL[\rvx]_{i+1} - \gL[\rvx]_{i}) \cdot \rva_{i}
\end{align}
where weight $\rvw_i = \frac{1}{2}(\gL[\rvx]_{i+1} + \gL[\rvx]_{i})$. 
We weigh the sum by distance to reflect the fact that outliers are separated in distance from the inliers. 
Points with both large distances to inliers and large average scores of the neighbors receive the largest weighted sum, which makes true outliers well separated from false positives with a larger weighted sum.

Note that we only compute the weighted sum for $\rvx$ based on all neighbors $\rvy$ such that $\|\rvx - \rvy \|_2 \leq r_{\max}$, instead of computing the sum under the entire clothes line. In other words, we only consider the average outlier scores of close neighbors of $\rvx$ instead of all points in $\gD$.
Here, $r_{\max}$ represents the distance from $\rvx$ to the nearest inlier point in $R_{\psi}$ (i.e. the projection of $\rvx$ to $R_{\psi}$) and is an estimation of the neighboring regions around $\rvx$ that should be focused. 
Setting $r_{\max}$ to be the maximum over all projections of $\rvx \in \gH$ to the inlier points gives us a sufficiently large radius that includes neighbors necessary to separate false positive points from true outliers.  In fact, the sum under the entire clothes line makes the difference of the weighted sums between false positives and true outliers less apparent (see Fig. \ref{fig:clothes_line_example}). Computing the sum up to $r_{\max}$ also makes the algorithm more efficient.

{\bf Deciding True Outlier Candidates.~}
We compute the weighted sum for both $\rvx \in \gH_{\rvc}$ and $\rvx \in \gR_{\psi}$, and build a candidate true outliers set as the points with a weighted sum that is greater than the average weighted sum $\text{mean}(\gA)$ (Line 19). The weighted sums of $\rvx \in \gR_{\psi}$, representing such sums of the inlier points, are used as comparison against that of the true outliers to make $\text{mean}(\gA)$ a meaningful threshold.

\subsubsection{Update the Neighbor Graph $\gG$}
The points in the candidate set $\gH_{c}$ can be seen as refined outlier micro-cluster representatives. Then, the neighbors of $\rvx \in \gH_{c}$ can be marked as points in the same outlier micro-cluster. To this end, we first estimate the diameter of the micro-cluster that includes $\rvx$, denoted $\tau_n$, based on $\gL[\rvx]$  (Line 22), and then compute the neighbors of $\rvx$, denoted $\gN$, as points $\rvy \in \gD$ within the diameter, i.e. $\|\rvy - \rvx\|_2 \leq \tau_n$ (Line 24). The diameter $\tau_n$ is found by a peak finding algorithm, \textsf{\small{detecta}}\footnote{\label{fn:detecta} detecta: \url{https://github.com/demotu/detecta}},
on the sorted distances $\gL[\rvx]$. Since outlier micro-clusters are separated in distance from the other points by definition, the first peak in the sorted distance $\gL[\rvx]$ indicates such a gap between the micro-cluster and the rest of the points in $\gD$. We store the information of the neighbors $\gN$ in graph $\gG$ by increasing the edge weight $(\rvx, \rvy)$ by one, $\forall \rvy \in \gN$ (Line 25).

\subsubsection{Pruning True Micro-cluster Representatives and Their Neighbors}
Finally, we prune the candidate true positive points $\rvx\in \gH_{c}$ along with their neighbors (Line 28). This effectively prunes the entire micro-cluster that contains $\rvx$, towards the goal of reducing ``masking''.

We remark that the synergy between micro-cluster assignment based on detection and leveraging micro-clusters to improve detection via pruning during a single iteration can be seen through the following perspectives: (1) We find outlier representatives $\gH$ based on outlier scores (i.e. current detection results); (2) The candidate set $\gH_{c}$ is computed based on both outlier representatives $\gH$ and neighboring outlier scores;
and (3) The candidate set $\gH_{c}$ is used for both pruning toward improving detection as well as outlier micro-cluster assignment -- where we find micro-clusters as neighbors of points $\rvx \in \gH_{c}$ and prune both $\rvx \in \gH_{c}$ and their neighbors. 

{\bf Outputting Outlier Micro-clusters Based on $\gG$.~}
At the end of our sequential ensemble, we find outlier micro-clusters based on the neighbor graph $\gG$. We first find a threshold $\tau_e$ to be the first peak of sorted edge weights (from the largest to the smallest) using \textsf{\small{detecta}}\textsuperscript{\ref{fn:detecta}}.
We keep only those strong edges with weights larger than $\tau_e$ to get a pruned graph $\gG'$. The outlier micro-clusters $\gC = \{\gC_1, \gC_2, \dots\}$ are output as the connected components of $\gG'$. See \textsf{\small{FindClusters}} in Appendix~\ref{sec:find_clusters}.

\subsection{Proposed \delac}
\label{subsec:delac}
Thus far, we have proposed \delaczero, a sequential ensemble that alternates between outlier detection and micro-cluster assignment. However, there exists a remaining key challenge: the choice of the hyperparameter $\psi$, i.e. the subsample size, for the base detector. Although \delaczero improves the robustness to $\psi$ relatively thanks to ``cleaning'' the dataset via outlier pruning, it remains sensitive to the choice (see e.g. Fig.~\ref{fig:sensitivity_to_psi} in Appendix~\ref{subsec:sensitivity_hyperparam}).
In prior works (e.g.~\cite{bandaragoda2018iNNE, ting2020idk}), only the best detection performance of iNNE with a chosen hyperparameter $\psi$ via a grid search is reported. However, $\psi$ is far from trivial to set in practice for fully \textit{unsupervised} settings. 

On one hand, $\psi$ cannot be smaller than the number of inlier clusters; otherwise, a small $\gR_{\psi}$ underrepresents the inlier distribution, which leads to many false positives as the hyperspheres constructed by a small $\gR_{\psi}$ are unable to cover the inliers effectively. On the other hand, $\psi$ cannot be too large. A large $\psi$ increases the chance of ``masking'' which leads to false negatives (micro-clusters) and hence overall poor detection. 
Therefore we ask the question:
{\em Can we make \delaczero even more robust to $\psi$?}

To this end, we propose \delac  (Algorithm~\ref{alg:delac}), where the main idea is to take advantage of \textit{hyper}-ensembling to increase hyperparameter robustness to specific settings \cite{ding2022hyperparameter}. 
As such, \delac consists of two phases: 
Phase 1 is a ``warm up'' phase (Lines 2-8), consisting of a hyper-ensemble of \delaczero models with varying values of $\psi \in [2,  \psi_{\max}]$, and Phase 2 simply employs the \delaczero algorithm one last time on the cleaned dataset using $\psi_{\max}$ (Line 12). 

As discussed earlier, a small $\psi$ is more likely to yield false positive errors, whereas a large $\psi$ is to yield false negatives due to ``masking''. We do not have a mechanism to recover from missed outliers, while the clothes-lines strategy can be applied to filter out some false positives. That is why in Phase 1 we vary $\psi$ $\in [2, \psi_{\max}]$, starting small and gradually increasing it.

The outcome of Phase 1 is a set of outlier micro-clusters by the hyper-ensemble (Line 9), which are pruned (Line 10) before \delaczero is employed in Phase 2, based on which the outlier micro-clusters are updated and the outlier scores are returned.
Note that Phase 2 uses a fixed subsample size of $\psi_{\max}$, i.e. the input and also the largest value that Phase 1 has used. Since Phase 2 trains \delaczero on the cleaned dataset, in the absence of most outlier micro-clusters, a larger subsample size poses a smaller risk for ``masking''.

\begin{algorithm}[!ht]
\caption{\textsc{\delac}}
\label{alg:delac}
\begin{algorithmic}[1]
\Require data $\gD$$\in$$\R^{n \times d}$, max subsample size $\psi_{\max}$, num. iterations $t$ (default: 100),  num. check-points $p$ (default: 0.1$n$)
\Ensure outlier scores $\bar{\rvs} \in [0, 1]^n$, explicit set of outlier micro-clusters $\gC=\{\gC_1, \gC_2, \ldots\}$
\State Neighbor graph $\gG \leftarrow (\gV = \gD, \gE = \emptyset)$ (init.)
\State \texttt{\# Phase 1: hyperensemble as } ``\texttt{warm up}''
\State $t^\prime \leftarrow \lfloor t/2 \rfloor$
\State $\Psi \leftarrow $ Pick $t^\prime$ values in $\in [2, \psi_{\max}]$ with equal gap
\For{$i = 1, 2, \dots, t^\prime$}
\State $\_, \_, \gG^{(i)} \leftarrow$ \textsc{\delaczero} ($\gD: \gD$, $\psi: \Psi[i]$, $t: i$, $p: p$)
\State $\gG \leftarrow$ increase weight of ($\rvx, \rvy$) by edge weight in $\gG^{(i)}$
\EndFor
\State $\gC_{\text{warmup}} \leftarrow $\textsf{\small{FindClusters}}($\gG$) \Comment{\texttt{\# Procedure~\ref{alg:find_clusters}}}
\State Clean set $\gR \leftarrow \gD \setminus \gC_{\text{warmup}}$
\State \texttt{\# Phase 2: sequential ensemble}
\State $\bar{\rvs}, \_, \gG_0 \leftarrow $ \textsc{\delaczero} ($\gD: \gR$, $\psi:\psi_{\max}$, $t:(t-t^\prime)$, $p:p$)
\State $\gG \leftarrow \gG \cup \gG_0$
\State $\gC \leftarrow $ \textsf{\small{FindClusters}}($\gG$)\\
\Return $\bar{\rvs}, \gC$
\end{algorithmic}
\end{algorithm}

\section{Experiments}
\label{sec:experiments}

\begin{table*}[!t]
 \caption{\textbf{\delac excels in micro-cluster assignment}: we report F1 scores of micro-cluster assignment for our methods \delac, and \delaczero, and also for $8$ state-of-the-art competitors. \delac outperforms everyone with the \textit{very best} average ranking; it is the winner or the runner up in $10$ out of our $16$ datasets. The \textbf{first} place is in bold, and \underline{second} place is underlined.}
    \label{tab:assignment_evaluation}
\begin{adjustbox}{width=2\columnwidth,center}
    \begin{tabular}{|c|cc|cc|
    cc|cc|cc|cc|cc
    |cc|c|}
    \hline
        \backslashbox{Dataset}{Method} 
        & \delaczero & \delac 
        & iNNE+O & iNNE+X 
        & LOF+O & LOF+X 
        & kNN+O & kNN+X 
        & iForest+O & iForest+X 
        & SCiForest+O & SCiForest+X 
        & COPOD+O & COPOD+X 
        & HBOS+O & HBOS+X 
        & \gensqout\\
    \hline
        \texttt{synthetic10} & \textbf{0.68} & \underline{0.67} & 0.41 & 0.35 & 0.32 & 0.31 & 0.49 & 0.29 & 0.43 & 0.61 & 0.53 & 0.46 & 0.25 & 0.41 & 0.39 & 0.47 & 0.29\\
        & \textbf{$\pm$ 0.47 (1)} & \underline{$\pm$ 0.47 (2)} & $\pm$ 0.21 (9) & $\pm$ 0.30 (12) & $\pm$ 0.25 (13) & $\pm$ 0.34 (14) & $\pm$ 0.15 (5) & $\pm$ 0.18 (15) & $\pm$ 0.20 (8) & $\pm$ 0.35 (3) & $\pm$ 0.13 (4) & $\pm$ 0.23 (7) & $\pm$ 0.26 (17) & $\pm$ 0.43 (10) & $\pm$ 0.20 (11) & $\pm$ 0.40 (6) & $\pm$ 0.44 (16)\\
        \texttt{spiral} & \underline{0.68} & \textbf{0.70} & 0.28 & 0.31 & 0.30 & 0.30 & 0.39 & 0.40 & 0.08 & 0.16 & 0.10 & 0.17 & 0.08 & 0.17 & 0.00 & 0.00 & 0.17\\
        & \underline{$\pm$ 0.47 (2)} & \textbf{$\pm$ 0.46 (1)} & $\pm$ 0.27 (8) & $\pm$ 0.38 (5) & $\pm$ 0.27 (7) & $\pm$ 0.35 (6) & $\pm$ 0.29 (4) & $\pm$ 0.38 (3) & $\pm$ 0.17 (14) & $\pm$ 0.36 (12) & $\pm$ 0.21 (13) & $\pm$ 0.37 (10) & $\pm$ 0.17 (14) & $\pm$ 0.37 (10) & $\pm$ 0.00 (16) & $\pm$ 0.00 (16) & $\pm$ 0.37 (10)\\
        \texttt{sandwich} & \textbf{0.97} & \underline{0.96} & 0.40 & 0.37 & 0.26 & 0.23 & 0.54 & 0.56 & 0.46 & 0.53 & 0.56 & 0.43 & 0.29 & 0.41 & 0.19 & 0.10 & 0.35\\
        & \textbf{$\pm$ 0.18 (1)} & \underline{$\pm$ 0.20 (2)} & $\pm$ 0.26 (10) & $\pm$ 0.30 (11) & $\pm$ 0.28 (14) & $\pm$ 0.30 (15) & $\pm$ 0.13 (5) & $\pm$ 0.32 (4) & $\pm$ 0.26 (7) & $\pm$ 0.39 (6) & $\pm$ 0.14 (3) & $\pm$ 0.16 (8) & $\pm$ 0.31 (13) & $\pm$ 0.42 (9) & $\pm$ 0.27 (16) & $\pm$ 0.16 (17) & $\pm$ 0.35 (12)\\
        \texttt{vdensity} & \textbf{0.94} & \underline{0.94} & 0.46 & 0.71 & 0.30 & 0.41  & 0.47 & 0.67 & 0.39 & 0.64 & 0.55 & 0.65 & 0.34 & 0.67 & 0.27 & 0.40 & 0.50\\
        & \textbf{$\pm$ 0.22 (1)} & \underline{$\pm$ 0.23 (2)} & $\pm$ 0.22 (11) & $\pm$ 0.39 (3) & $\pm$ 0.27 (16) & $\pm$ 0.42 (12) & $\pm$ 0.23 (10) & $\pm$ 0.41 (5) & $\pm$ 0.24 (14) & $\pm$ 0.42 (7) & $\pm$ 0.09 (8) & $\pm$ 0.27 (6) & $\pm$ 0.25 (15) & $\pm$ 0.47 (4) & $\pm$ 0.27 (17) & $\pm$ 0.44 (13) & $\pm$ 0.50 (9)\\
        \hline
        \texttt{letter} & \underline{0.88} & \textbf{0.98} & 0.36 & 0.46 & 0.26 & 0.34 & 0.42 & 0.46 & 0.63 & 0.52 & 0.63 & 0.79 & 0.63 & 0.76 & 0.63 & 0.42 & 0.00\\
        & \underline{$\pm$ 0.33 (2)} & \textbf{$\pm$ 0.15 (1)} & $\pm$ 0.34 (14) & $\pm$ 0.44 (11) & $\pm$ 0.34 (16) & $\pm$ 0.42 (15) & $\pm$ 0.37 (12) & $\pm$ 0.42 (10) & $\pm$ 0.26 (6) & $\pm$ 0.30 (9) & $\pm$ 0.26 (6) & $\pm$ 0.32 (3) & $\pm$ 0.26 (6) & $\pm$ 0.32 (4) & $\pm$ 0.26 (6) & $\pm$ 0.25 (13) & $\pm$ 0.00 (17)\\
        \texttt{musk} & \underline{0.96} & \textbf{0.99} & 0.52 & 0.46 & 0.45 & 0.42 & 0.48 & 0.46 & 0.71 & 0.60 & 0.71 & 0.66 & 0.71 & 0.67 & 0.71 & 0.53 & 0.00\\
        & \underline{$\pm$ 0.19 (2)} & \textbf{$\pm$ 0.10 (1)} & $\pm$ 0.39 (11) & $\pm$ 0.41 (14) & $\pm$ 0.42 (15) & $\pm$ 0.44 (16) & $\pm$ 0.42 (12) & $\pm$ 0.44 (13) & $\pm$ 0.31 (4) & $\pm$ 0.36 (9) & $\pm$ 0.31 (4) & $\pm$ 0.38 (8) & $\pm$ 0.31 (4) & $\pm$ 0.37 (7) & $\pm$ 0.31 (4) & $\pm$ 0.33 (10) & $\pm$ 0.00 (17)\\
        \texttt{thyroid} & 0.92 & \underline{0.96} & 0.27 & 0.35 & 0.10 & 0.16 & 0.41 & 0.48 & 0.62 & 0.70 & 0.62 & 0.77 & 0.62 & 0.83 & 0.62 & 0.87 & \textbf{1.00}\\
        & $\pm$ 0.27 (3) & \underline{$\pm$ 0.20 (2)} & $\pm$ 0.32 (15) & $\pm$ 0.43 (14) & $\pm$ 0.21 (17) & $\pm$ 0.34 (16) & $\pm$ 0.32 (13) & $\pm$ 0.43 (12) & $\pm$ 0.15 (9) & $\pm$ 0.32 (7) & $\pm$ 0.15 (9) & $\pm$ 0.28 (6) & $\pm$ 0.15 (9) & $\pm$ 0.26 (5) & $\pm$ 0.15 (9) & $\pm$ 0.21 (4) & \textbf{$\pm$ 0.00 (1)}\\
        \texttt{optdigits} & \underline{0.88} & \textbf{0.96} & 0.31 & 0.30 & 0.24 & 0.22 & 0.35 & 0.34 & 0.54 & 0.56 & 0.54 & 0.54 & 0.53 & 0.39 & 0.52 & 0.48 & 0.00\\
        & \underline{$\pm$ 0.33 (2)} & \textbf{$\pm$ 0.19 (1)} & $\pm$ 0.31 (13) & $\pm$ 0.39 (14) & $\pm$ 0.30 (15) & $\pm$ 0.35 (16) & $\pm$ 0.31 (11) & $\pm$ 0.40 (12) & $\pm$ 0.23 (6) & $\pm$ 0.40 (3) & $\pm$ 0.23 (4) & $\pm$ 0.39 (5) & $\pm$ 0.22 (7) & $\pm$ 0.36 (10) & $\pm$ 0.23 (8) & $\pm$ 0.39 (9) & $\pm$ 0.00 (17)\\
        \texttt{satimage-2} & 0.45 & \textbf{0.82} & 0.31 & 0.44 & 0.34 & 0.47 & 0.41 & 0.58 & 0.61 & 0.68 & 0.61 & \underline{0.70} & 0.37 & 0.28 & 0.11 & 0.11 & 0.00\\
        & $\pm$ 0.50 (8) & \textbf{$\pm$ 0.39 (1)} & $\pm$ 0.35 (13) & $\pm$ 0.47 (9) & $\pm$ 0.37 (12) & $\pm$ 0.47 (7) & $\pm$ 0.36 (10) & $\pm$ 0.45 (6) & $\pm$ 0.27 (5) & $\pm$ 0.32 (3) & $\pm$ 0.27 (4) & \underline{$\pm$ 0.29 (2)} & $\pm$ 0.35 (11) & $\pm$ 0.34 (14) & $\pm$ 0.18 (15) & $\pm$ 0.23 (16) & $\pm$ 0.00 (17)\\
\hline
        \texttt{lympho} & 0.00 & 0.00 & 0.48 & 0.50 & 0.55 & 0.57 & 0.41 & 0.43 & 0.40 & 0.44 & 0.49 & \underline{0.60} & 0.53 & \textbf{0.64} & 0.38 & 0.36 & 0.00\\
        & $\pm$ 0.00 (16) & $\pm$ 0.00 (16) & $\pm$ 0.27 (8) & $\pm$ 0.27 (6) & $\pm$ 0.30 (4) & $\pm$ 0.23 (3) & $\pm$ 0.32 (11) & $\pm$ 0.33 (10) & $\pm$ 0.35 (12) & $\pm$ 0.36 (9) & $\pm$ 0.37 (7) & \underline{$\pm$ 0.35 (2)} & $\pm$ 0.28 (5) & \textbf{$\pm$ 0.25 (1)} & $\pm$ 0.38 (13) & $\pm$ 0.36 (14) & $\pm$ 0.00 (16)\\
        \texttt{ecoli} & \underline{0.42} & 0.38 & 0.26 & 0.25 & 0.31 & 0.26 & 0.38 & 0.36 & 0.40 & 0.33 & \textbf{0.46} & 0.38 & 0.22 & 0.20 & 0.30 & 0.24 & 0.00\\
        & \underline{$\pm$ 0.42 (2)} & $\pm$ 0.41 (4) & $\pm$ 0.29 (11) & $\pm$ 0.29 (13) & $\pm$ 0.27 (9) & $\pm$ 0.23 (12) & $\pm$ 0.35 (6) & $\pm$ 0.33 (7) & $\pm$ 0.31 (3) & $\pm$ 0.24 (8) & \textbf{$\pm$ 0.36 (1)} & $\pm$ 0.30 (5) & $\pm$ 0.17 (15) & $\pm$ 0.16 (16) & $\pm$ 0.24 (10) & $\pm$ 0.20 (14) & $\pm$ 0.00 (17)\\
        \texttt{musk\_real} & 0.47 & 0.49 & 0.37 & 0.55 & 0.21 & 0.34 & 0.32 & 0.49 & 0.53 & \textbf{0.87} & 0.52 & \textbf{0.87} & 0.07 & 0.07 & 0.40 & 0.65 & 0.00\\
        & $\pm$ 0.17 (9) & $\pm$ 0.20 (8) & $\pm$ 0.24 (11) & $\pm$ 0.39 (4) & $\pm$ 0.23 (14) & $\pm$ 0.41 (12) & $\pm$ 0.25 (13) & $\pm$ 0.41 (7) & $\pm$ 0.05 (5) & \textbf{$\pm$ 0.13 (1)} & $\pm$ 0.05 (6) & \textbf{$\pm$ 0.13 (1)} & $\pm$ 0.07 (15) & $\pm$ 0.08 (16) & $\pm$ 0.06 (10) & $\pm$ 0.08 (3) & $\pm$ 0.00 (17)\\
        \texttt{satellite} & \textbf{0.59} & \underline{0.57} & 0.13 & 0.12 & 0.28 & 0.16 & 0.24 & 0.20 & 0.29 & 0.33 & 0.27 & 0.33 & 0.27 & 0.33 & 0.27 & 0.33 & 0.00\\
        & \textbf{$\pm$ 0.49 (1)} & \underline{$\pm$ 0.49 (2)} & $\pm$ 0.27 (15) & $\pm$ 0.30 (16) & $\pm$ 0.43 (8) & $\pm$ 0.27 (14) & $\pm$ 0.38 (12) & $\pm$ 0.37 (13) & $\pm$ 0.37 (7) & $\pm$ 0.47 (4) & $\pm$ 0.39 (10) & $\pm$ 0.47 (4) & $\pm$ 0.39 (10) & $\pm$ 0.47 (4) & $\pm$ 0.39 (10) & $\pm$ 0.47 (4) & $\pm$ 0.00 (17)\\
        \texttt{shuttle} & 0.14 & 0.21 & 0.40 & 0.25 & 0.37 & 0.24 & 0.55 & 0.34 & 0.60 & \underline{0.64} & \textbf{0.65} & 0.59 & 0.45 & 0.53 & 0.22 & 0.32 & 0.00\\
        & $\pm$ 0.35 (16) & $\pm$ 0.40 (15) & $\pm$ 0.45 (8) & $\pm$ 0.32 (12) & $\pm$ 0.44 (9) & $\pm$ 0.31 (13) & $\pm$ 0.44 (5) & $\pm$ 0.34 (10) & $\pm$ 0.23 (3) & \underline{$\pm$ 0.31 (2)} & \textbf{$\pm$ 0.40 (1)} & $\pm$ 0.37 (4) & $\pm$ 0.32 (7) & $\pm$ 0.33 (6) & $\pm$ 0.22 (14) & $\pm$ 0.37 (11) & $\pm$ 0.00 (17)\\
        \texttt{smtp} & 0.05 & 0.06 & 0.05 & 0.04 & 0.09 & 0.09 & 0.03 & 0.03 & 0.04 & 0.03 & 0.03 & 0.03 & 0.04 & 0.03 & \textbf{0.25} & \underline{0.23} & 0.23\\
        & $\pm$ 0.13 (8) & $\pm$ 0.16 (6) & $\pm$ 0.09 (7) & $\pm$ 0.07 (10) & $\pm$ 0.15 (5) & $\pm$ 0.19 (4) & $\pm$ 0.05 (16) & $\pm$ 0.05 (13) & $\pm$ 0.07 (9) & $\pm$ 0.05 (13) & $\pm$ 0.05 (17) & $\pm$ 0.05 (13) & $\pm$ 0.06 (10) & $\pm$ 0.05 (13) & \textbf{$\pm$ 0.21 (1)} & \underline{$\pm$ 0.24 (2)} & $\pm$ 0.23 (3)\\
        \texttt{http} & 0.00 & 0.08 & 0.22 & 0.15 & 0.22 & 0.16 & 0.12 & 0.12 & 0.19 & 0.15 & \underline{0.25} & 0.23 & \textbf{0.34} & 0.21 & 0.18 & 0.10 & 0.17\\
        & $\pm$ 0.06 (17) & $\pm$ 0.27 (16) & $\pm$ 0.33 (5) & $\pm$ 0.30 (12) & $\pm$ 0.31 (4) & $\pm$ 0.29 (10) & $\pm$ 0.30 (14) & $\pm$ 0.30 (13) & $\pm$ 0.36 (7) & $\pm$ 0.34 (11) & \underline{$\pm$ 0.35 (2)} & $\pm$ 0.33 (3) & \textbf{$\pm$ 0.37 (1)} & $\pm$ 0.32 (6) & $\pm$ 0.37 (8) & $\pm$ 0.26 (15) & $\pm$ 0.37 (9)\\
\hline
Avg. Rank & 5.69 & \textbf{5.00} & 10.56 & 10.38 & 11.12 & 11.56 & 9.94 & 9.56 & 7.44 & 6.69 & 6.19 & \underline{5.44} & 9.94 & 8.44 & 10.50 & 10.44 & 13.25\\
    \hline
    \end{tabular}
\end{adjustbox}
\end{table*}

\begin{table*}[!t]
  \caption{\textbf{\delac is  competitive in detection}: we report Average Precision (AP) scores for $10$ detectors.
  Our \delac obtains the second best average ranking.
  Note that the best detector \gensqout is actually the \textit{worst} method in assignment; therefore, we have a much better balance w.r.t. these two tasks.
  The \textbf{first} place is in bold, and \underline{second} place is underlined.}
    \label{tab:detection_evaluation_ap}
\begin{adjustbox}{width=2\columnwidth,center}
    \begin{tabular}{|c|cc|c|c|c|c|c|c|c|c|}
    \hline
        \backslashbox{Dataset}{Method} 
        & \delaczero & \delac & iNNE & iForest & SCiForest & LOF & kNN & 
        COPOD & HBOS & \gensqout\\
    \hline
    \texttt{synthetic10} & 0.86 $\pm$ 0.17 (4) & \underline{0.88 $\pm$ 0.14 (2)} & 0.73 $\pm$ 0.29 (7) & 0.71 $\pm$ 0.27 (8) & 0.87 $\pm$ 0.29 (3) & 0.57 $\pm$ 0.39 (9) & 0.79 $\pm$ 0.30 (5) & 
    0.53 $\pm$ 0.00 (10) & 0.76 $\pm$ 0.00 (6) & \textbf{0.95 $\pm$ 0.00 (1)}\\
    \texttt{spiral} & \underline{0.58 $\pm$ 0.37 (2)} & 0.57 $\pm$ 0.38 (3) & 0.46 $\pm$ 0.38 (5) & 0.16 $\pm$ 0.08 (8) & 0.17 $\pm$ 0.06 (7) & 0.57 $\pm$ 0.44 (3) & \textbf{0.68 $\pm$ 0.45 (1)} & 
    0.03 $\pm$ 0.00 (10) & 0.04 $\pm$ 0.00 (9) & 0.22 $\pm$ 0.01 (6)\\
    \texttt{sandwich} & \underline{0.98 $\pm$ 0.04 (2)} & \textbf{1.00 $\pm$ 0.01 (1)} & 0.67 $\pm$ 0.36 (6) & 0.51 $\pm$ 0.31 (7) & 0.81 $\pm$ 0.34 (5) & 0.43 $\pm$ 0.36 (8) & 0.84 $\pm$ 0.24 (4) & 
    0.36 $\pm$ 0.00 (9) & 0.27 $\pm$ 0.00 (10) & 0.90 $\pm$ 0.01 (3)\\
    \texttt{vdensity} & \underline{0.98 $\pm$ 0.07 (2)} & \underline{0.98 $\pm$ 0.07 (2)} & 0.79 $\pm$ 0.30 (5) & 0.59 $\pm$ 0.29 (8) & 0.83 $\pm$ 0.28 (4) & 0.47 $\pm$ 0.39 (10) & 0.77 $\pm$ 0.33 (6) & 
    0.69 $\pm$ 0.00 (7) & 0.52 $\pm$ 0.00 (9) & \textbf{1.00 $\pm$ 0.00 (1)}\\
    \hline
    \texttt{letter} & 0.89 $\pm$ 0.20 (5) & 0.93 $\pm$ 0.11 (4) & 0.64 $\pm$ 0.36 (9) & 0.88 $\pm$ 0.31 (6) & 0.88 $\pm$ 0.31 (6) & 0.51 $\pm$ 0.34 (10) & 0.71 $\pm$ 0.41 (8) & 
    \textbf{1.00 $\pm$ 0.00 (1)} & \textbf{1.00 $\pm$ 0.00 (1)} & \textbf{1.00 $\pm$ 0.00 (1)}\\
    \texttt{musk} & 0.93 $\pm$ 0.14 (4) & 0.93 $\pm$ 0.12 (4) & 0.75 $\pm$ 0.34 (8) & 0.88 $\pm$ 0.31 (7) & 0.89 $\pm$ 0.30 (6) & 0.66 $\pm$ 0.40 (10) & 0.70 $\pm$ 0.42 (9) & 
    \textbf{1.00 $\pm$ 0.00 (1)} & \textbf{1.00 $\pm$ 0.00 (1)} & \textbf{1.00 $\pm$ 0.00 (1)}\\
    \texttt{thyroid} & 0.92 $\pm$ 0.11 (5) & 0.94 $\pm$ 0.09 (4) & 0.48 $\pm$ 0.38 (9) & 0.90 $\pm$ 0.29 (6) & 0.90 $\pm$ 0.29 (6) & 0.26 $\pm$ 0.26 (10) & 0.73 $\pm$ 0.38 (8) & 
    \textbf{1.00 $\pm$ 0.00 (1)} & \textbf{1.00 $\pm$ 0.00 (1)} & \textbf{1.00 $\pm$ 0.00 (1)}\\
    \texttt{optdigits} & 0.88 $\pm$ 0.19 (7) & 0.94 $\pm$ 0.10 (4) & 0.60 $\pm$ 0.39 (9) & 0.90 $\pm$ 0.29 (5) & 0.90 $\pm$ 0.29 (5) & 0.48 $\pm$ 0.34 (10) & 0.70 $\pm$ 0.42 (8) & 
    \textbf{1.00 $\pm$ 0.00 (1)} & \textbf{1.00 $\pm$ 0.00 (1)} & \textbf{1.00 $\pm$ 0.00 (1)}\\
    \texttt{satimage-2} & \underline{0.95 $\pm$ 0.06 (2)} & 0.93 $\pm$ 0.09 (3) & 0.57 $\pm$ 0.36 (8) & 0.58 $\pm$ 0.38 (6) & 0.88 $\pm$ 0.29 (4) & 0.58 $\pm$ 0.42 (6) & 0.70 $\pm$ 0.42 (5) & 
    0.53 $\pm$ 0.00 (9) & 0.33 $\pm$ 0.00 (10) & \textbf{0.97 $\pm$ 0.01 (1)}\\
    \hline
    \texttt{lympho} & 0.67 $\pm$ 0.14 (6) & 0.67 $\pm$ 0.15 (6) & 0.63 $\pm$ 0.16 (8) & 0.54 $\pm$ 0.31 (9) & 0.72 $\pm$ 0.35 (3) & 0.68 $\pm$ 0.25 (5) & 0.53 $\pm$ 0.10 (10) & 
    \textbf{0.88 $\pm$ 0.00 (1)} & \underline{0.86 $\pm$ 0.00 (2)} & 0.72 $\pm$ 0.08 (3)\\
    \texttt{ecoli} & 0.54 $\pm$ 0.20 (3) & \underline{0.55 $\pm$ 0.21 (2)} & 0.40 $\pm$ 0.26 (5) & 0.18 $\pm$ 0.19 (10) & 0.39 $\pm$ 0.31 (6) & 0.39 $\pm$ 0.18 (6) & \textbf{0.67 $\pm$ 0.17 (1)} & 
    0.25 $\pm$ 0.00 (9) & 0.44 $\pm$ 0.00 (4) & 0.29 $\pm$ 0.05 (8)\\
    \texttt{musk\_real} & 0.78 $\pm$ 0.37 (3) & 0.76 $\pm$ 0.39 (4) & 0.71 $\pm$ 0.42 (5) & 0.33 $\pm$ 0.42 (9) & 0.65 $\pm$ 0.37 (6) & 0.37 $\pm$ 0.45 (8) & 0.60 $\pm$ 0.44 (7) & 
    0.06 $\pm$ 0.00 (10) & \underline{0.88 $\pm$ 0.00 (2)} & \textbf{1.00 $\pm$ 0.00 (1)}\\
    \texttt{satellite} & 0.66 $\pm$ 0.28 (6) & 0.69 $\pm$ 0.23 (5) & 0.35 $\pm$ 0.34 (9) & 0.61 $\pm$ 0.32 (7) & 0.73 $\pm$ 0.24 (4) & 0.22 $\pm$ 0.29 (10) & 0.42 $\pm$ 0.40 (8) & 
    0.77 $\pm$ 0.00 (3) & \textbf{0.82 $\pm$ 0.00 (1)} & \underline{0.81 $\pm$ 0.01 (2)}\\
    \texttt{shuttle} & 0.36 $\pm$ 0.16 (6) & 0.36 $\pm$ 0.16 (6) & 0.34 $\pm$ 0.20 (8) & 0.62 $\pm$ 0.32 (4) & \underline{0.76 $\pm$ 0.27 (2)} & 0.20 $\pm$ 0.15 (10) & 0.31 $\pm$ 0.18 (9) & 
    0.53 $\pm$ 0.00 (5) & 0.72 $\pm$ 0.00 (3) & \textbf{0.79 $\pm$ 0.04 (1)}\\
    \texttt{smtp} & 0.12 $\pm$ 0.10 (4) & 0.12 $\pm$ 0.09 (4) & 0.13 $\pm$ 0.09 (3) & 0.08 $\pm$ 0.06 (8) & 0.12 $\pm$ 0.04 (4) & 0.12 $\pm$ 0.14 (4) & 0.04 $\pm$ 0.02 (10) & 
    0.07 $\pm$ 0.00 (9) & \textbf{0.33 $\pm$ 0.00 (1)} & \underline{0.17 $\pm$ 0.00 (2)}\\
    \texttt{http} & 0.14 $\pm$ 0.05 (7) & 0.15 $\pm$ 0.05 (6) & 0.16 $\pm$ 0.05 (4) & 0.10 $\pm$ 0.07 (9) & \textbf{0.20 $\pm$ 0.09 (1)} & \textbf{0.20 $\pm$ 0.11 (1)} & 0.08 $\pm$ 0.05 (10) & 
    0.11 $\pm$ 0.00 (8) & 0.19 $\pm$ 0.00 (3) & 0.16 $\pm$ 0.02 (4)\\
    \hline
    Avg. Rank & 4.25 & \underline{3.75} & 6.75 & 7.31 & 4.50 & 7.50 & 6.81 & 
    5.87 & 4.00 & \textbf{2.31}\\
    \hline
    \end{tabular}
\end{adjustbox}
  \vspace{-0.05in}
\end{table*}

We evaluate \delac against 16 datasets and 8 SOTA detectors and 2 hyperparameter-free clustering algorithms w.r.t. both detection and micro-cluster assignment. We also provide an in-depth analysis of its various elements in the Appendix.

{\bf Datasets.~}
Our datasets fall into three main categories:
\begin{enumerate*}
\item \textit{2D synthetic datasets};
\item \textit{Semi-synthetic datasets}: Datasets \texttt{letter}, \texttt{musk}, \texttt{thyroid}, \texttt{optdigits}, and \texttt{satimage-2}
\item \textit{Real-world benchmark datasets}: Datasets \texttt{lympho}, \texttt{ecoli}, \texttt{musk\_real}, \texttt{satellite}, \texttt{shuttle}, \texttt{smtp}, and \texttt{http}.
\end{enumerate*}
Both \textit{Semi-synthetic datasets} and \textit{Real-world benchmark datasets} come from the ODDS repository\footnote{\url{http://odds.cs.stonybrook.edu}} with preprocessing that ensures the datasets contain small, compact outlier micro-clusters. See Appendix~\ref{ssec:datasets_appendix} for more details.

{\bf Baselines and Configurations.~}We evaluate both tasks, detection and micro-cluster assignment, with separate experiments; accordingly, we consider two different sets of baselines: {\em Detection baselines} and {\em Assignment baselines}. See Appendix~\ref{subsec:baselines_appendix} and~\ref{subsec:config_appendix} for details on baselines and configurations used in the experiments.

{\bf Performance Metrics.~} 
\begin{enumerate*}
    \item \textit{Detection}: We report both the area under the ROC curve (ROC AUC) and the Average Precision (AP).
    \item \textit{Assignment}: We report the average F1 score against the true set of outlier micro-clusters.
\end{enumerate*}
For methods with hyperparameters, we report the \textit{averaged} performance and one stdev over the list of hyperparameter settings and 5 random runs each. For hyperparameter-free algorithms, we report the average performance and one stdev only over 5 random runs.

\subsection{Assignment Evaluation}

Here we evaluate the methods w.r.t. outlier micro-cluster assignment.
\delac, and \delaczero do assignment in-house.
\gensqout has a post-hoc assignment procedure.
For the other methods, we post-process the outliers they detect using two of-the-shelf clustering algorithms: 
X-means, and OPTICS.
They are both parameter-free, and well-regarded in the literature.
Table~\ref{tab:assignment_evaluation} reports the results of this experiment.
We use ``+O'' to indicate post-hoc clustering with OPTICS; ``+X'' stands for X-means.
Note that our proposed methods are notably effective in assignment.
\delac, and \delaczero obtain respectively the best and the third best average rankings among all datasets.
Also, \delac is the best or the second best method in $10$ out of our $16$ datasets; \delaczero does the same for $9$ datasets.
Finally, note that \gensqout has the worst average ranking among all $17$ methods tested; interestingly, (to the best of our knowledge) it is the only method in literature that is originally designed to perform micro-cluster assignment.

\subsection{Detection Evaluation}

The results of detection in AP scores are in Table~\ref{tab:detection_evaluation_ap}. See Appendix~\ref{subsec:detection_eval_auc} for results in AUC scores.
As it can be seen, our methods are quite competitive in terms of detection.
Note that \delac and \delaczero are respectively the second and the fourth best methods in the average ranking of AP, while a similar pattern is observed in the AUC scores.
The best performing method is \gensqout.
It is quite surprising as \gensqout is the worst performing method w.r.t. assignment (Table~\ref{tab:assignment_evaluation}). 
A similar scenario is also seen for HBOS, which is both one of the top performing detectors, yet one of the worst performing in assignment.
These results make it clear that our proposal is undoubtedly the best option among all the $10$ methods studied; \delac and \delaczero are the only methods that excel both in detection \textit{as well as} in assignment.


{\bf Acknowledgments~}{This work is partly sponsored by the PwC Risk and Regulatory Services Innovation Center at Carnegie Mellon University, and by the Brazilian foundations CAPES (finance code 001), FAPESP (grants 2021/05623-2, 2020/07200-9, 2018/05714-5 and 2016/17078-0), and CNPq.
}

\bibliographystyle{IEEEtran}
\bibliography{references}

\newpage
\clearpage
\section*{\Large Appendix}
\label{sec:appendix}
\setcounter{section}{0}

\section{Preliminaries}

\subsection{Notations}
\label{subsec:notations}
Let $\gD = [\rvx_1, \dots, \rvx_n]^T \in \R^{n \times d}$ be the input point-cloud dataset with $n$ $d$-dimensional instances. Let the outliers be $\gO = \gS \cup \{\gC_i\}_{i=1}^{m} \subseteq \gD$, where $\gS$ denotes the scattered singleton  outliers and $\gC_i \subseteq \gD, \forall i\in[m]$ denote $m$ outlier micro-clusters. Furthermore, $\gC_i \cap \gC_j = \emptyset, \forall i \neq j$ and $\gC_i \cap \gS = \emptyset, \forall i\in [m]$. $\|\cdot\|_2$ denotes the Euclidean distance. For a set $\gA$ (a vector $\rva$), we use $\gA[\rvx]$ ($\rva[\rvx]$) to denote the value in the set (vector) associated with a specific instance $\rvx$, and $\rva_i$ to denote the value of $\rva$ at index $i$. We use superscript as in $\gA^{(i)}$ or $\rva^{(i)}$ to denote a variable at the $i$-th iteration.

\subsection{iNNE Outlier Detector}
\label{subsec:inne_algo}

\delac uses individual models of Isolation using Nearest Neighbor Ensemble (iNNE)~\cite{bandaragoda2018iNNE} as the base outlier detector. In a nutshell, iNNE subsamples at random $\psi$ training points (centers or representatives), constructs a hypersphere centered at each instance in the sample, and 
assigns outlier scores to points in the test set based on how likely a point is covered by the hyperspheres. Since outliers are far from the majority by definition, they are less likely to be covered and thus receive high outlier scores. For completeness, we give the pseudocode of iNNE in Algorithm \ref{alg:inne_single}.

\begin{algorithm}
\caption{iNNE~\cite{bandaragoda2018iNNE}}
\label{alg:inne_single}
{
\begin{algorithmic}[1]
\Require training data $\gX$, test data $\gD$, subsample size $\psi$
\Ensure outlier scores $\rvs$, subsample $\gR_{\psi}$
\State $\gR_{\psi} \leftarrow$ Subsample $\psi$ points $\in \gX$ without replacement
\State Construct hyperspheres $\gH_{\rvc}$ centered at $\rvc \in R_{\psi}$ with radius $\text{rad}(\gH_{\rvc}) = \min_{\rvy \in \gR_{\psi}, \rvy \neq \rvc}\|\rvc - \rvy\|_2$, $\forall \rvc \in \gR_{\psi}$
\For{$\rvx \in \gD$}
\State \texttt{\# Assign outlier scores}
\If{$\rvx \notin \gH_{\rvc}$ for any $\gH_{\rvc}$}
\State $s(\rvx) \leftarrow 1$
\Else
\State $\gH_{\rvb} \leftarrow \argmin_{\text{rad}(\gH_{\rvb})} \{\gH_{\rvb}: \rvx \in \gH_{\rvb}\}$
\State $\rva \leftarrow \min_{\rva \in R_{\psi}, \;\rva \neq \rvb} \; \|\rva - \rvb\|_2$
\State $s(\rvx) \leftarrow 1 - \text{rad}(\gH_{\rva}) / \text{rad}(\gH_{\rvb})$
\EndIf
\EndFor
\State \Return $\rvs, \gR_{\psi}$
\end{algorithmic}
}
\end{algorithm}

\section{\textsc{FindClusters} Subroutine}
\label{sec:find_clusters}

\newfloat{subroutine}{H}{loa}
\floatname{subroutine}{Procedure}
\makeatletter\newcommand\l@subroutine{\@dottedtocline{1}{1.5em}{2.3em}}\makeatother
\begin{subroutine}[!h]
\caption{\textsf{\small{FindClusters}}}
\label{alg:find_clusters}
\begin{algorithmic}[1]
\Require neighbor graph $\gG$
\Ensure outlier micro-clusters $\gC$
\State $W \leftarrow$ sorted edge weights $\in \gG$ from largest to smallest
\State threshold $\tau_{e} \leftarrow$ \textsf{\small{FindFirstPeak}}($W$)
\State Pruned graph $\gG^\prime \leftarrow $ $\{\text{edge}(\rvx, \rvy) \in \gG: \text{weight}(\rvx, \rvy) > \tau_e\}$\\
\Return $\gC \leftarrow$ Connected components of $\gG^\prime$
\end{algorithmic}
\end{subroutine}

\section{Experimental Setup}
Experiments are set up to answer the following questions:
\cbit
\item Compared with state-of-the-art (SOTA) methods, how accurate is our \delac in micro-cluster assignment?
\item Compared with SOTA methods, how accurate is \delac in detecting outliers?
\item How sensitive to hyperparameter $\psi$ is \delac? And, how sensitive is its baseline method iNNE?
\item How much masking can \delac avoid?
\ceit

\subsection{Datasets}
\label{ssec:datasets_appendix}

\begin{figure}
    \centering
    \begin{subfigure}[b]{0.23\linewidth}   
        \centering 
        \includegraphics[width=\linewidth]{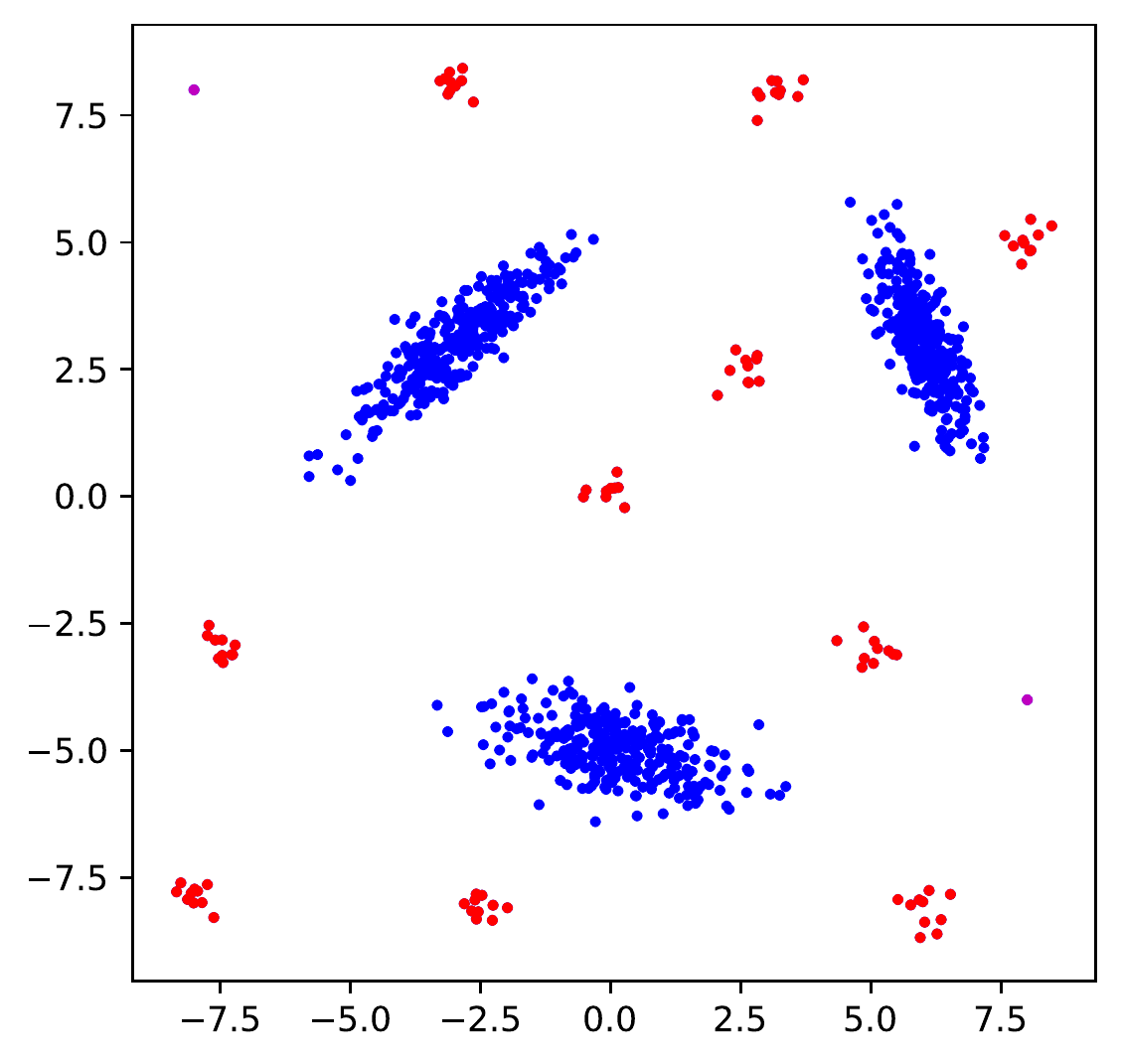}
        \caption{{\small \texttt{synth10}}}
    \end{subfigure}
    \begin{subfigure}[b]{0.23\linewidth}  
        \centering 
        \includegraphics[width=\linewidth]{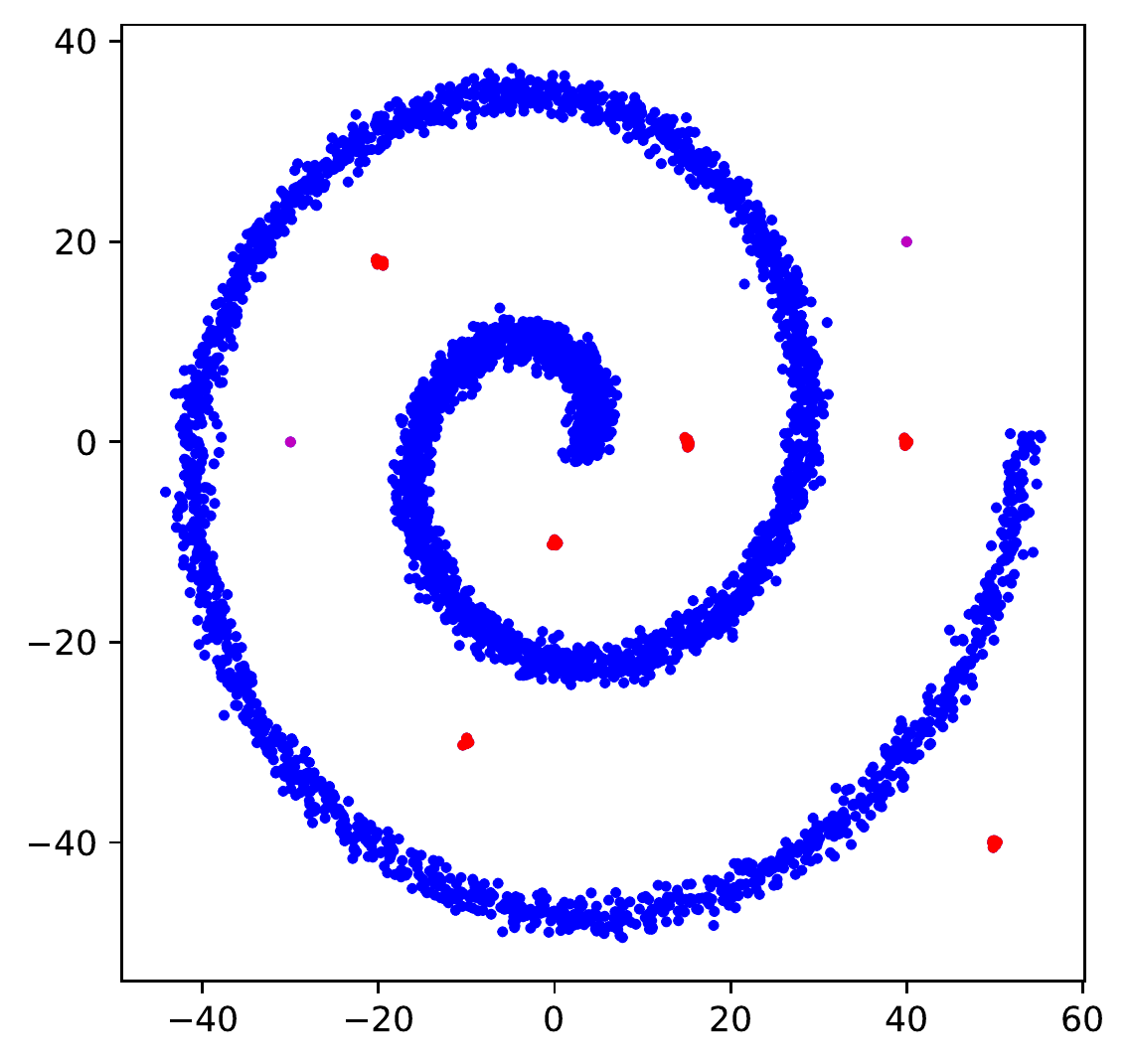}
        \caption{{\small \texttt{spiral}}}
    \end{subfigure}
    \begin{subfigure}[b]{0.23\linewidth}
        \centering
        \includegraphics[width=\linewidth]{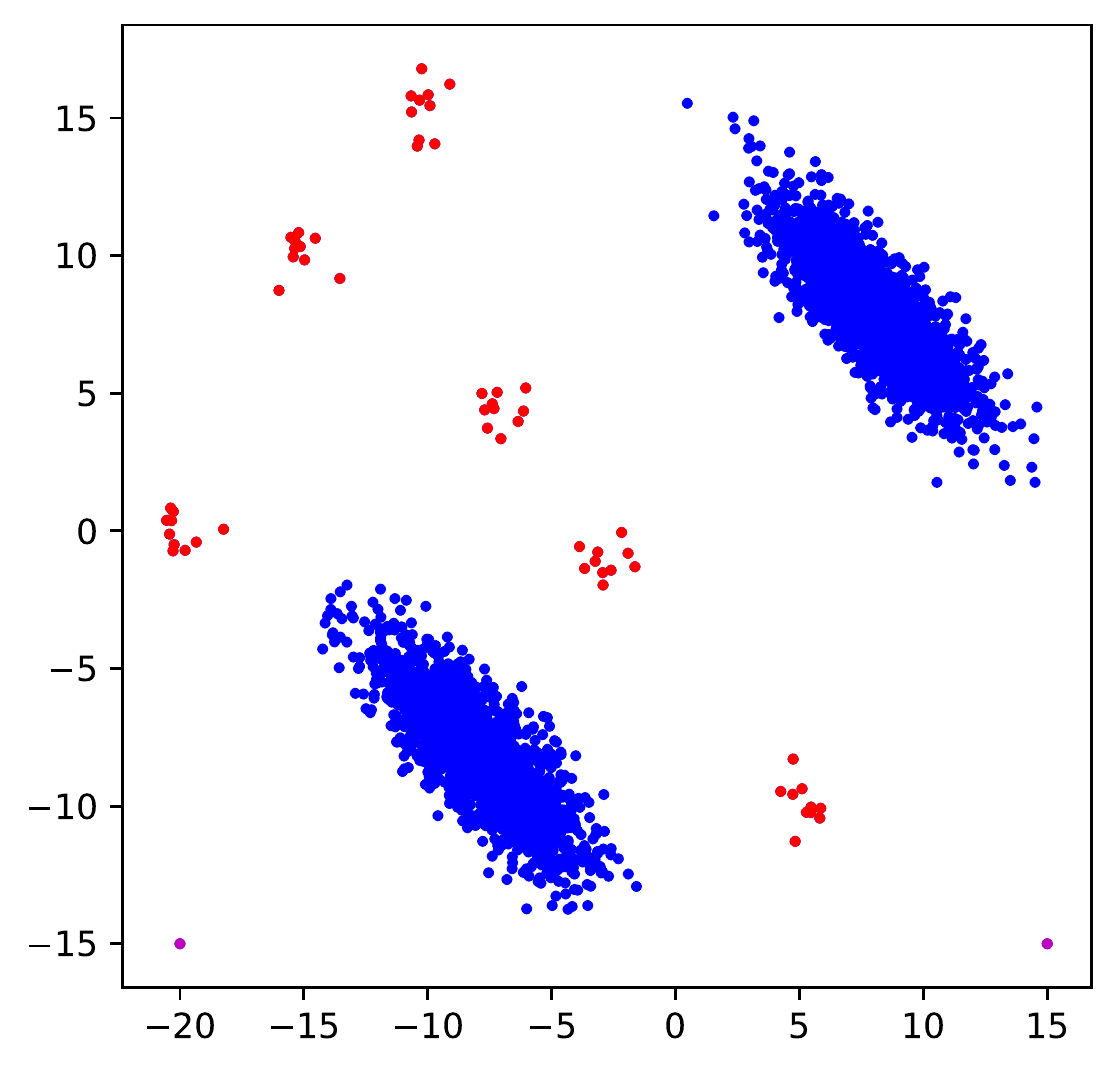}
        \caption{{\small \texttt{sandwich}}}
    \end{subfigure}
    \begin{subfigure}[b]{0.23\linewidth}   
        \centering 
        \includegraphics[width=\linewidth]{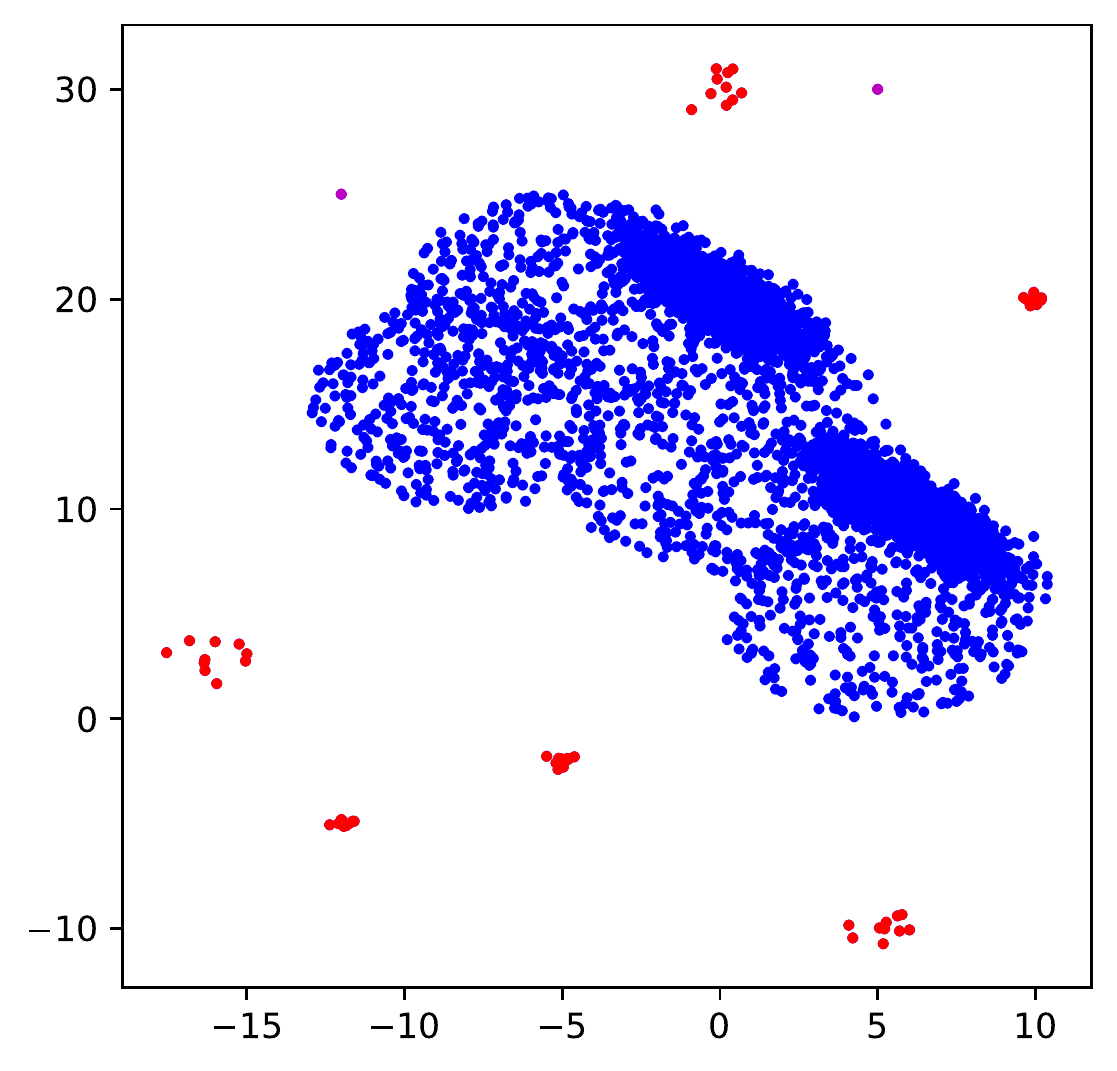}
        \caption{{\small \texttt{vdensity}}}
    \end{subfigure}
    \caption{Datasets of the `2D synthetic' category.}
    \label{fig:synthetic_datasets}
\end{figure}

Our datasets are listed in Table~\ref{tab:summary_dataset}\footnote{All datasets used in the experiments are available \href{https://drive.google.com/drive/folders/1GIqMDtVjGpYicZEtP1T06AuZc2b8hpPF?usp=sharing}{here}.}. 
They fall into three main categories:
\begin{enumerate}
\item \textit{2D synthetic datasets}: See datasets in Fig.~\ref{fig:synthetic_datasets};
\item \textit{Semi-synthetic datasets}: Datasets \texttt{letter}, \texttt{musk}, \texttt{thyroid}, \texttt{optdigits}, and \texttt{satimage-2} were randomly selected from the ODDS repository\footnote{\url{http://odds.cs.stonybrook.edu}}.
We first excluded the real outliers;
then, we simulated outliers from the real inliers by following the approach of Zhang et al.~(2022)~\citeapp{sparx22}.
Specifically, a Gaussian mixture model was fitted on the real inliers; and, outlying micro-clusters were generated from Gaussians with distinct means and distinct standard deviations.
\item \textit{Real-world benchmark datasets}: Datasets \texttt{lympho}, \texttt{ecoli}, \texttt{musk\_real}, \texttt{satellite}, \texttt{shuttle}, \texttt{smtp}, and \texttt{http} also come from ODDS.
These datasets were originally conceived for multi-class classification; latter, they were adapted for outlier detection by treating as outliers two or more small classes.
As micro-clusters of outliers must be compact and small, we further down-sampled the outliers by randomly picking one outlier from each class, and combining it with its $k \in [5, 30]$ nearest neighbors to form a micro-cluster.
\end{enumerate}

\begin{table}[H]
    \caption{A summary of the datasets.}
    \label{tab:summary_dataset}
\begin{adjustbox}{width=\columnwidth,center}
    \begin{tabular}{|c|c|c|c|c|c|c|}
    \hline
        Dataset & $n$ & $d$ & \# Outliers (\%)  & \# Micro-clusters & \# Samples per Micro-cluster \\
    \hline
        \texttt{synthetic10} & 1,002 & 2 & 102 (10.18 \%) & 10 & \{ 10 \} * 10 \\
        \texttt{spiral} & 4,062 & 2 &  62 (1.53 \%) &  6 & \{ 10 \} * 6 \\
        \texttt{sandwich} & 6,062 & 2  & 62 (1.02 \%) & 6 & \{ 10 \} * 6 \\
        \texttt{vdensity} & 6,062  & 2 & 62 (1.02 \%)  & 6 & \{ 10 \} * 6 \\
    \hline
        \texttt{letter} & 1,600 & 32 & 100 (6.25\%)  & 9 & \{ 10 \} * 9\\
        \texttt{musk} & 3,064 & 166 & 99 (3.23\%) & 9 & \{ 10 \} * 9\\
        \texttt{thyroid} & 3,778 & 6 & 99 (2.62\%) & 9 & \{ 10 \} * 9 \\
        \texttt{optdigits} & 5,221 & 64 & 155 (2.96\%) & 14 & \{ 10 \} * 14\\
        \texttt{satimage-2} & 5,809 & 36 & 77 (1.33\%) & 7 & \{ 10 \} * 7 \\
    \hline
        \texttt{lympho} & 148 & 18 & 6 (4.05 \%) & 2 & \{ 2, 4 \} \\
        \texttt{ecoli} & 316 & 7 & 9 (2.85 \%) & 3 & \{ 2, 2, 5 \} \\
        \texttt{musk\_real} & 2,985 & 166 & 20 (0.67 \%) & 2 & \{ 10, 10 \} \\
        \texttt{satellite} & 4,425 & 36 & 26 (0.59 \%) & 3 & \{ 20, 3, 3 \} \\
        \texttt{shuttle} & 5,050 & 9 & 50 (0.99 \%) & 5 & \{ 5, 5, 30, 5, 5 \} \\
        \texttt{smtp} & 5,053 & 3 & 53 (1.05 \%) & 4 & \{ 13, 10, 15, 14 \} \\
        \texttt{http} & 5,061 & 3 & 61 (1.21 \%) & 6 & \{ 10, 13, 10, 4, 16, 7 \} \\
    \hline
    \end{tabular}
\end{adjustbox}
\end{table}

\subsection{Baselines}
\label{subsec:baselines_appendix}
We evaluate both tasks, detection and micro-cluster assignment, with separate experiments; accordingly, we consider two different sets of baselines.

\begin{enumerate}
    \item \textit{Detection baselines}: 
    \begin{enumerate*}
        \item Subsampling based (subsample size $\psi$ is common hyperparameter): iNNE~\citeapp{bandaragoda2018iNNE}, iForest~\citeapp{liu2008isolation_forest}, and SCiForest~\citeapp{liu2010SCiForest};
        \item Nearest neighbor based (\# neighbors $k$ is common hyperparameter): kNN~\citeapp{ramaswamy2000efficient}, and LOF~\citeapp{breunig2000LOF};
        \item Hyperparameter-free: COPOD~\citeapp{li2020copod}, HBOS~\citeapp{Goldstein2012HBOS}\footnote{\# bins is automatically decided based on the Birge-Rozenblac method, which is optimal according to~\citeapp{Birge2006opt_bin_number}.}, \gensqout~\citeapp{lee2021gen2out}\footnote{We use the full configuration as suggested in~\citeapp{lee2021gen2out}.}.
    \end{enumerate*}
    \item \textit{Assignment baselines}:
    \begin{enumerate*}
        \item Two-stage: Detection baselines above (except \gensqout) paired with hyperparameter-free clustering algorithms \{X-Means~\citeapp{pelleg2000x}, OPTICS~\citeapp{ankerst1999optics}\} applied to top $m$ points with largest outliers scores, $m$ being the true \# outliers; 
        \item SOTA baseline: \gensqout.
    \end{enumerate*}
\end{enumerate}

\subsection{Configurations}
\label{subsec:config_appendix}
The subsample size $\psi$ and the number of models in the ensemble $t$ are two hyperparameters for the proposed \delaczero, \delac and subsampling based baselines. We follow prior works to set $t = 100$~\cite{liu2008isolation_forest, liu2010SCiForest, bandaragoda2018iNNE} and pick $\psi \in \{2, 4, 8, 16, \dots, \min(1024, 30\%n)\}$, where $n$ is the dataset size. The number of neighbors $k$ is a hyperparameter for nearest neighbor based baselines, where we use $k \in \{1, 5, 10, 20, 30, 50\}$. Note that such heuristics for picking $\psi$ or $k$ are widely adopted in prior works, e.g.~\cite{breunig2000LOF, bandaragoda2018iNNE, ting2020idk}.

\section{Additional Results}
\label{subsec:detection_eval_auc}
\subsection{Detection Evaluation in AUC scores}

The results of detection in AUC scores are in Table~\ref{tab:detection_evaluation_auc}.

\begin{table*}[!t]
 \caption{\textbf{\delac is  competitive in detection}: we report ROC-AUC scores for $10$ detectors.
  \delac obtains the third best average ranking.
  Note that the two best detectors \gensqout, and HBOS are actually among the \textit{worst} methods in assignment; therefore, we have a much better balance w.r.t. these two tasks.
  The \textbf{first} place is in bold, and \underline{second} place is underlined.}
    \label{tab:detection_evaluation_auc}
\begin{adjustbox}{width=2\columnwidth,center}
    \begin{tabular}{|c|cc|c|c|c|c|c|c|c|c|}
    \hline
        \backslashbox{Dataset}{Method} 
        & \delaczero & \delac & iNNE & iForest & SCiForest & LOF & kNN & 
        COPOD & HBOS & \gensqout\\
    \hline
    \texttt{synthetic10} & \underline{0.95 $\pm$ 0.08 (2)} & \underline{0.95 $\pm$ 0.08 (2)} & 0.91 $\pm$ 0.14 (7) & 0.86 $\pm$ 0.16 (8) & 0.93 $\pm$ 0.16 (6) & 0.79 $\pm$ 0.23 (9) & \underline{0.95 $\pm$ 0.08 (2)} & 
    0.72 $\pm$ 0.00 (10) & \underline{0.95 $\pm$ 0.00 (2)} & \textbf{0.99 $\pm$ 0.00 (1)}\\
    \texttt{spiral} & \textbf{0.86 $\pm$ 0.19 (1)} & \textbf{0.86 $\pm$ 0.19 (1)} & 0.82 $\pm$ 0.20 (4) & 0.61 $\pm$ 0.11 (9) & 0.63 $\pm$ 0.10 (8) & 0.84 $\pm$ 0.22 (3) & 0.73 $\pm$ 0.39 (6) & 
    0.55 $\pm$ 0.00 (10) & 0.70 $\pm$ 0.00 (7) & 0.74 $\pm$ 0.02 (5)\\
    \texttt{sandwich} & \textbf{1.00 $\pm$ 0.01 (1)} & \textbf{1.00 $\pm$ 0.00 (1)} & 0.95 $\pm$ 0.11 (5) & 0.87 $\pm$ 0.19 (8) & 0.94 $\pm$ 0.15 (6) & 0.76 $\pm$ 0.20 (9) & \textbf{1.00 $\pm$ 0.01 (1)} & 
    0.66 $\pm$ 0.00 (10) & 0.89 $\pm$ 0.00 (7) & \textbf{1.00 $\pm$ 0.00 (1)}\\
    \texttt{vdensity} & \textbf{1.00 $\pm$ 0.00 (1)} & \textbf{1.00 $\pm$ 0.00 (1)} & 0.98 $\pm$ 0.04 (4) & 0.90 $\pm$ 0.14 (8) & 0.95 $\pm$ 0.15 (6) & 0.76 $\pm$ 0.22 (10) & 0.95 $\pm$ 0.08 (6) & 
    0.97 $\pm$ 0.00 (5) & 0.90 $\pm$ 0.00 (8) & \textbf{1.00 $\pm$ 0.00 (1)}\\
    \hline
    \texttt{letter} & 0.97 $\pm$ 0.06 (5) & 0.99 $\pm$ 0.02 (4) & 0.91 $\pm$ 0.12 (8) & 0.94 $\pm$ 0.17 (6) & 0.94 $\pm$ 0.17 (6) & 0.65 $\pm$ 0.26 (10) & 0.70 $\pm$ 0.42 (9) & 
    \textbf{1.00 $\pm$ 0.00 (1)} & \textbf{1.00 $\pm$ 0.00 (1)} & \textbf{1.00 $\pm$ 0.00 (1)}\\
    \texttt{musk} & 0.99 $\pm$ 0.03 (4) & 0.99 $\pm$ 0.02 (4) & 0.96 $\pm$ 0.09 (6) & 0.94 $\pm$ 0.16 (7) & 0.94 $\pm$ 0.16 (7) & 0.74 $\pm$ 0.31 (9) & 0.70 $\pm$ 0.43 (10) & 
    \textbf{1.00 $\pm$ 0.00 (1)} & \textbf{1.00 $\pm$ 0.00 (1)} & \textbf{1.00 $\pm$ 0.00 (1)}\\
    \texttt{thyroid} & 0.98 $\pm$ 0.04 (5) & 0.99 $\pm$ 0.04 (4) & 0.90 $\pm$ 0.14 (9) & 0.95 $\pm$ 0.15 (7) & 0.95 $\pm$ 0.15 (7) & 0.65 $\pm$ 0.22 (10) & 0.97 $\pm$ 0.05 (6) & 
    \textbf{1.00 $\pm$ 0.00 (1)} & \textbf{1.00 $\pm$ 0.00 (1)} & \textbf{1.00 $\pm$ 0.00 (1)}\\
    \texttt{optdigits} & 0.96 $\pm$ 0.08 (5) & 0.99 $\pm$ 0.03 (4) & 0.91 $\pm$ 0.14 (8) & 0.95 $\pm$ 0.15 (6) & 0.95 $\pm$ 0.15 (6) & 0.65 $\pm$ 0.28 (10) & 0.70 $\pm$ 0.43 (9) & 
    \textbf{1.00 $\pm$ 0.00 (1)} & \textbf{1.00 $\pm$ 0.00 (1)} & \textbf{1.00 $\pm$ 0.00 (1)}\\
    \texttt{satimage-2} & \textbf{1.00 $\pm$ 0.01 (1}) & \textbf{1.00 $\pm$ 0.00 (1)} & 0.95 $\pm$ 0.08 (6) & 0.93 $\pm$ 0.15 (8) & 0.95 $\pm$ 0.15 (6) & 0.77 $\pm$ 0.29 (10) & 0.86 $\pm$ 0.20 (9) & 
    0.99 $\pm$ 0.00 (4) & 0.99 $\pm$ 0.00 (4) & \textbf{1.00 $\pm$ 0.00 (1)}\\
    \hline
    \texttt{lympho} & 0.98 $\pm$ 0.01 (3) & 0.98 $\pm$ 0.01 (3) & 0.97 $\pm$ 0.02 (6) & 0.87 $\pm$ 0.19 (10) & 0.89 $\pm$ 0.20 (9) & 0.95 $\pm$ 0.09 (8) & 0.97 $\pm$ 0.01 (6) & 
    \textbf{0.99 $\pm$ 0.00 (1)} & \textbf{0.99 $\pm$ 0.00 (1)} & 0.98 $\pm$ 0.01 (3)\\
    \texttt{ecoli} & \textbf{0.87 $\pm$ 0.03 (1)} & \textbf{0.87 $\pm$ 0.04 (1)} & 0.85 $\pm$ 0.07 (4) & 0.69 $\pm$ 0.16 (10) & 0.77 $\pm$ 0.16 (8) & 0.75 $\pm$ 0.17 (9) & \textbf{0.87 $\pm$ 0.05 (1)} & 
    0.79 $\pm$ 0.00 (7) & 0.80 $\pm$ 0.00 (6) & 0.82 $\pm$ 0.01 (5)\\
    \texttt{musk\_real} & 0.98 $\pm$ 0.07 (3) & 0.96 $\pm$ 0.10 (4) & 0.95 $\pm$ 0.11 (5) & 0.80 $\pm$ 0.21 (9) & 0.94 $\pm$ 0.16 (6) & 0.69 $\pm$ 0.27 (10) & 0.81 $\pm$ 0.33 (8) & 
    0.91 $\pm$ 0.00 (7) & \textbf{1.00 $\pm$ 0.00 (1)} & \textbf{1.00 $\pm$ 0.00 (1)}\\
    \texttt{satellite} & 0.89 $\pm$ 0.11 (6) & 0.90 $\pm$ 0.12 (5) & 0.82 $\pm$ 0.17 (8) & 0.87 $\pm$ 0.14 (7) & 0.91 $\pm$ 0.14 (3) & 0.58 $\pm$ 0.26 (10) & 0.69 $\pm$ 0.27 (9) & 
    0.91 $\pm$ 0.00 (3) & \textbf{0.99 $\pm$ 0.00 (1)} & \underline{0.95 $\pm$ 0.00 (2)}\\
    \texttt{shuttle} & 0.92 $\pm$ 0.12 (6) & 0.92 $\pm$ 0.12 (6) & 0.92 $\pm$ 0.12 (6) & 0.93 $\pm$ 0.15 (5) & 0.95 $\pm$ 0.15 (4) & 0.71 $\pm$ 0.19 (9) & 0.66 $\pm$ 0.24 (10) & 
    \underline{0.99 $\pm$ 0.00 (2)} & 0.97 $\pm$ 0.00 (3) & \textbf{1.00 $\pm$ 0.00 (1)}\\
    \texttt{smtp} & 0.73 $\pm$ 0.23 (8) & 0.74 $\pm$ 0.24 (6) & 0.74 $\pm$ 0.24 (6) & 0.82 $\pm$ 0.13 (5) & 0.88 $\pm$ 0.13 (3) & 0.48 $\pm$ 0.12 (9) & 0.47 $\pm$ 0.22 (10) & 
    0.87 $\pm$ 0.00 (4) & \textbf{0.97 $\pm$ 0.00 (1)} & \underline{0.94 $\pm$ 0.00 (2)}\\
    \texttt{http} & 0.44 $\pm$ 0.14 (7) & 0.43 $\pm$ 0.13 (9) & 0.44 $\pm$ 0.14 (7) & \underline{0.86 $\pm$ 0.12 (2)} & \textbf{0.88 $\pm$ 0.13 (1)} & 0.51 $\pm$ 0.10 (6) & 0.25 $\pm$ 0.08 (10) & 
    0.67 $\pm$ 0.00 (4) & 0.57 $\pm$ 0.00 (5) & 0.83 $\pm$ 0.01 (3)\\
    \hline
    Avg. Rank & 3.68 & 3.50 & 6.18 & 7.18 & 5.75 & 8.81 & 7.00 & 
    4.43 & \underline{3.12} & \textbf{1.87}\\
    \hline
    \end{tabular}
\end{adjustbox}
 
\end{table*}

\subsection{Sensitivity to hyperparameter $\psi$}
\label{subsec:sensitivity_hyperparam}

\begin{figure}[!th]
    \centering
    \begin{subfigure}[b]{0.48\linewidth}
        \centering
        \includegraphics[width=\linewidth]{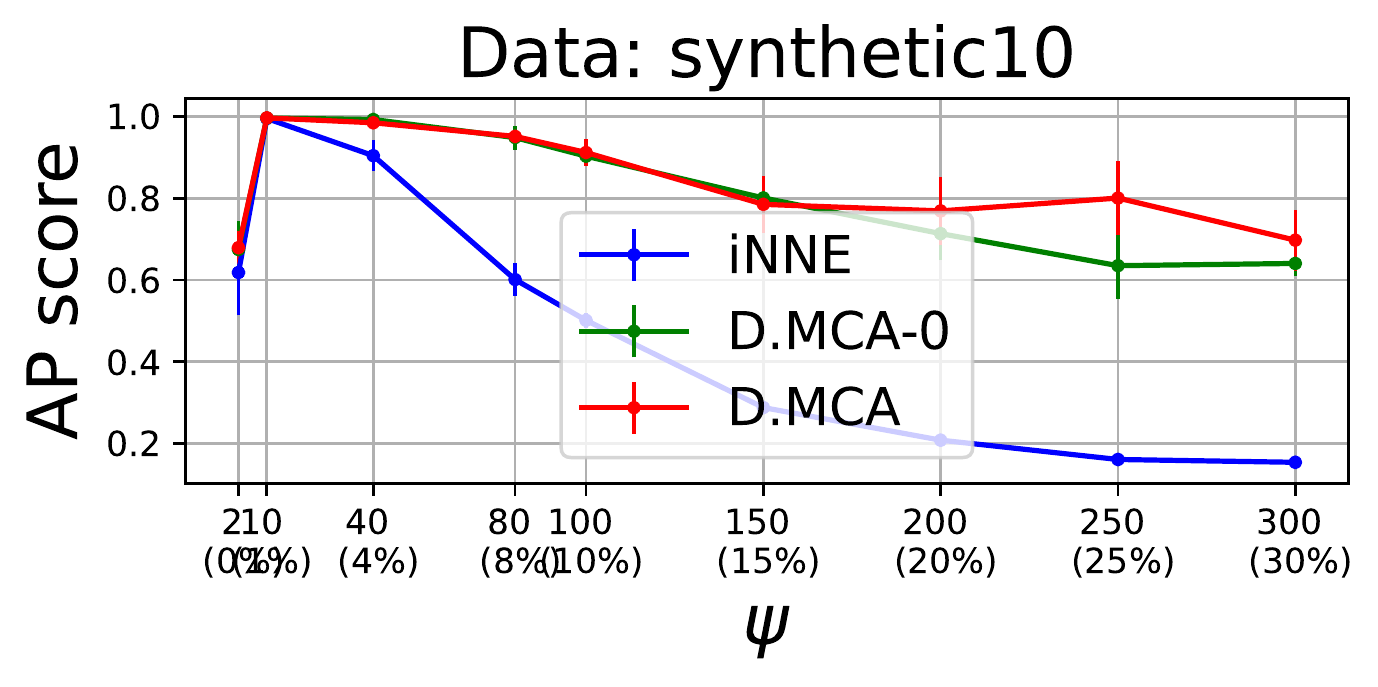}
    \end{subfigure}
    \hfill
    \begin{subfigure}[b]{0.48\linewidth}  
        \centering 
        \includegraphics[width=\linewidth]{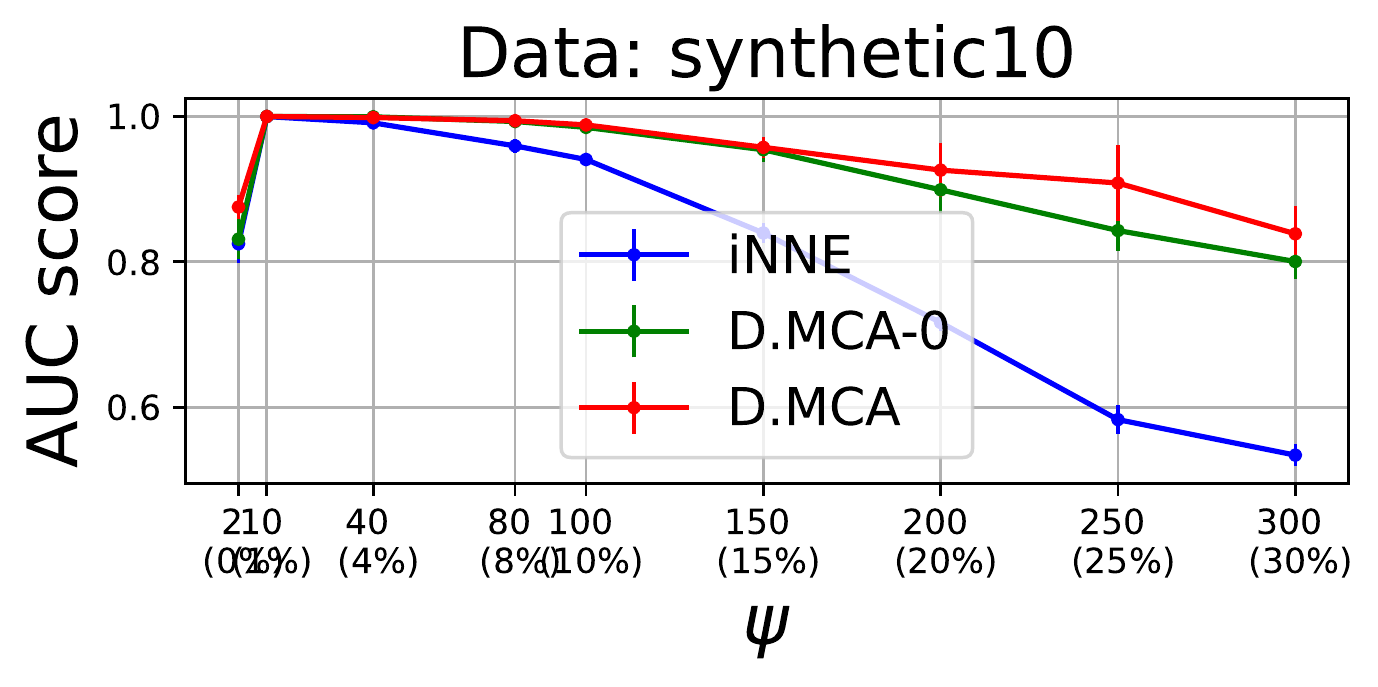}
    \end{subfigure}
    \begin{subfigure}[b]{0.475\linewidth}   
        \centering 
        \includegraphics[width=\linewidth]{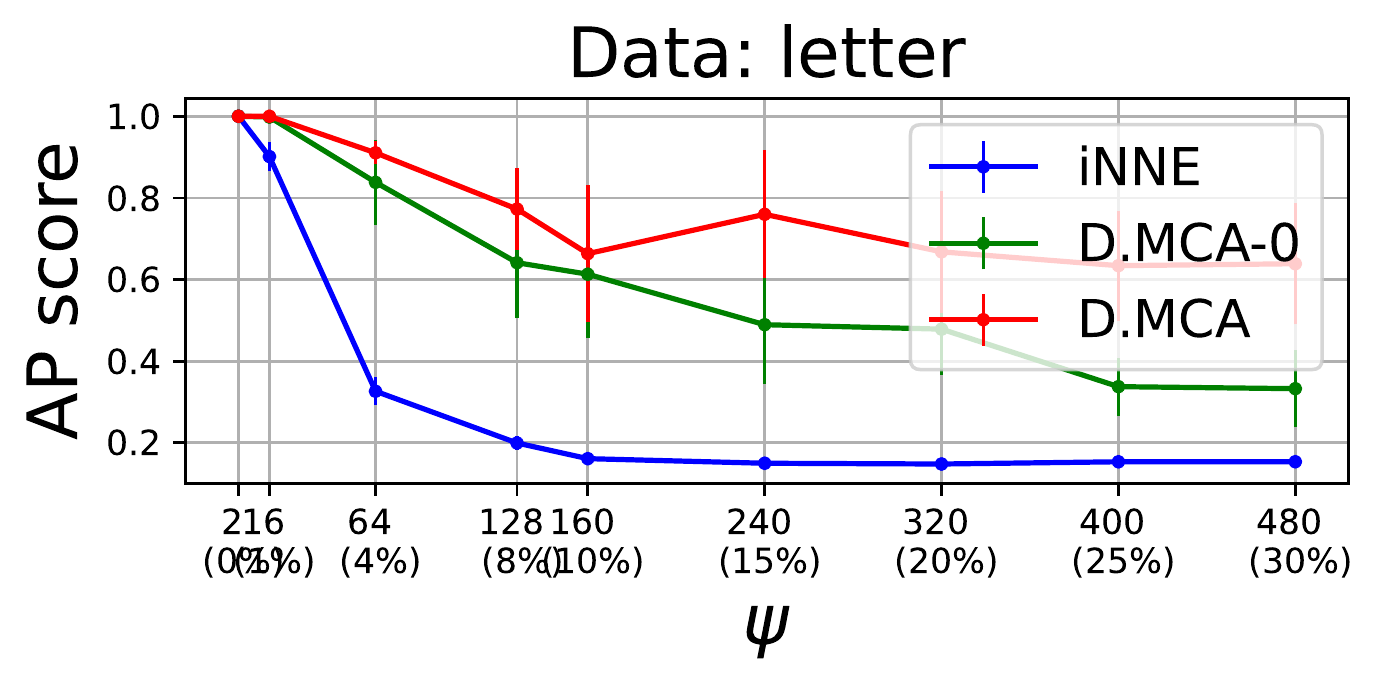}
    \end{subfigure}
    \hfill
    \begin{subfigure}[b]{0.475\linewidth}   
        \centering 
        \includegraphics[width=\linewidth]{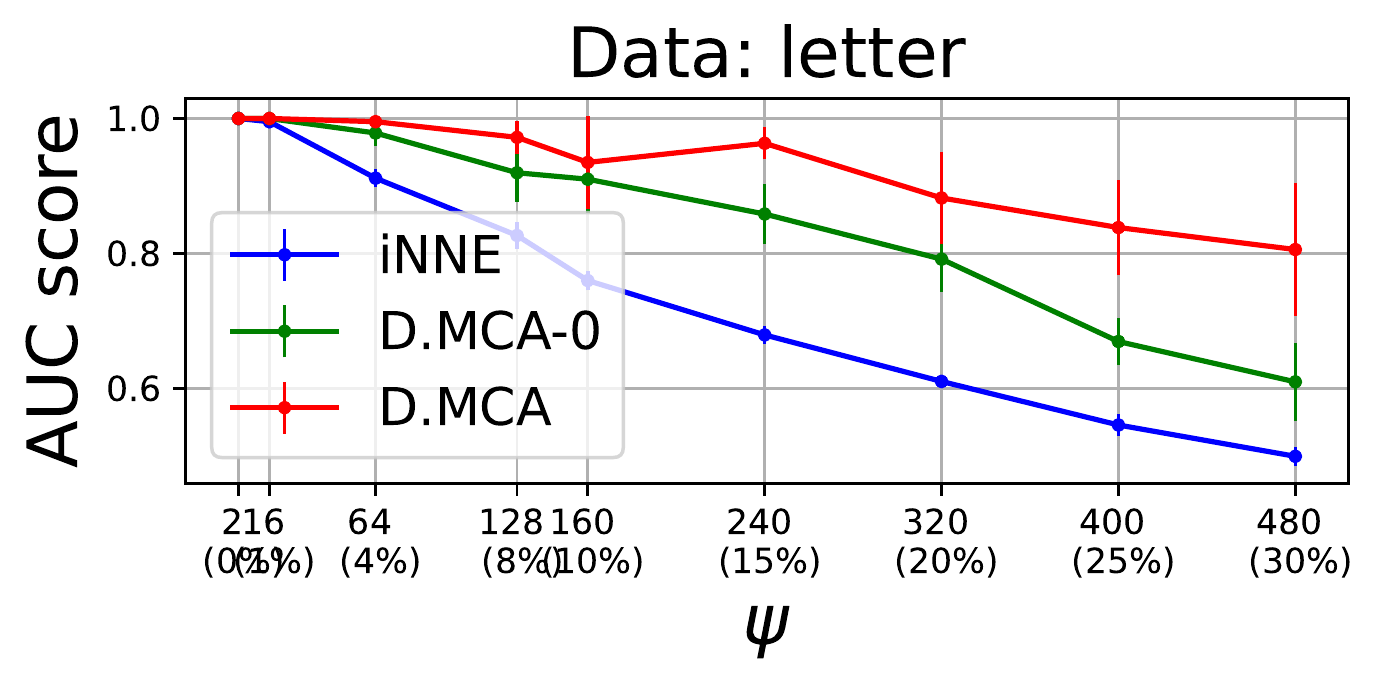}
    \end{subfigure}
    \vspace{-0.05in}
    \caption{A comparison on the sensitivity to $\psi$ between \textcolor{blue}{iNNE}, \textcolor{green}{\delaczero} and \textcolor{red}{\delac} on \texttt{synthetic10} and \texttt{letter}.}
    \label{fig:sensitivity_to_psi}
\end{figure}

Here we evaluate the sensitivity of \delac, \delaczero, and iNNE to hyperparameter~$\psi$.
Fig.~\ref{fig:sensitivity_to_psi} (top) reports the AP (left) and AUC (right) scores obtained from dataset \texttt{synthetic10}.
As it is expected, the three methods perform poorly when using $\psi=2$; this is because a subsample size of $2$ is not large enough to represent all $3$ clusters of inliers in this dataset; see the blue clusters in Fig~\ref{fig:synthetic_datasets}(a).
Using $\psi=10$ leads therefore to much better results; this configuration allows all methods to return near-perfect scores.
The scores tend to decrease due to masking as $\psi$ increases further; however, thanks to our hyper-ensemble and pruning strategies, both \delac and \delaczero are considerably less sensitive than iNNE to large $\psi$.
As $\psi$ increases, the scores of iNNE degrade much faster than those of our methods, with \delac being even more robust to poor choices of $\psi$ than \delaczero.
We obtained similar results from our other datasets; the results for dataset~\texttt{letter} are shown in  Fig.~\ref{fig:sensitivity_to_psi} (bottom).
This particular dataset has $2$ clusters of inliers -- see the PCA $2$D projection in the right side of Fig.~\ref{fig:clothes_lines_2d_data_synthetic10}; thus, all methods work well with $\psi=2$.
The results for the other datasets are omitted for brevity.

\subsection{Outlier Pruning \& Masking}

\begin{figure}[!t]
    \begin{subfigure}[b]{0.475\linewidth}
        \centering 
        \includegraphics[width=\linewidth]{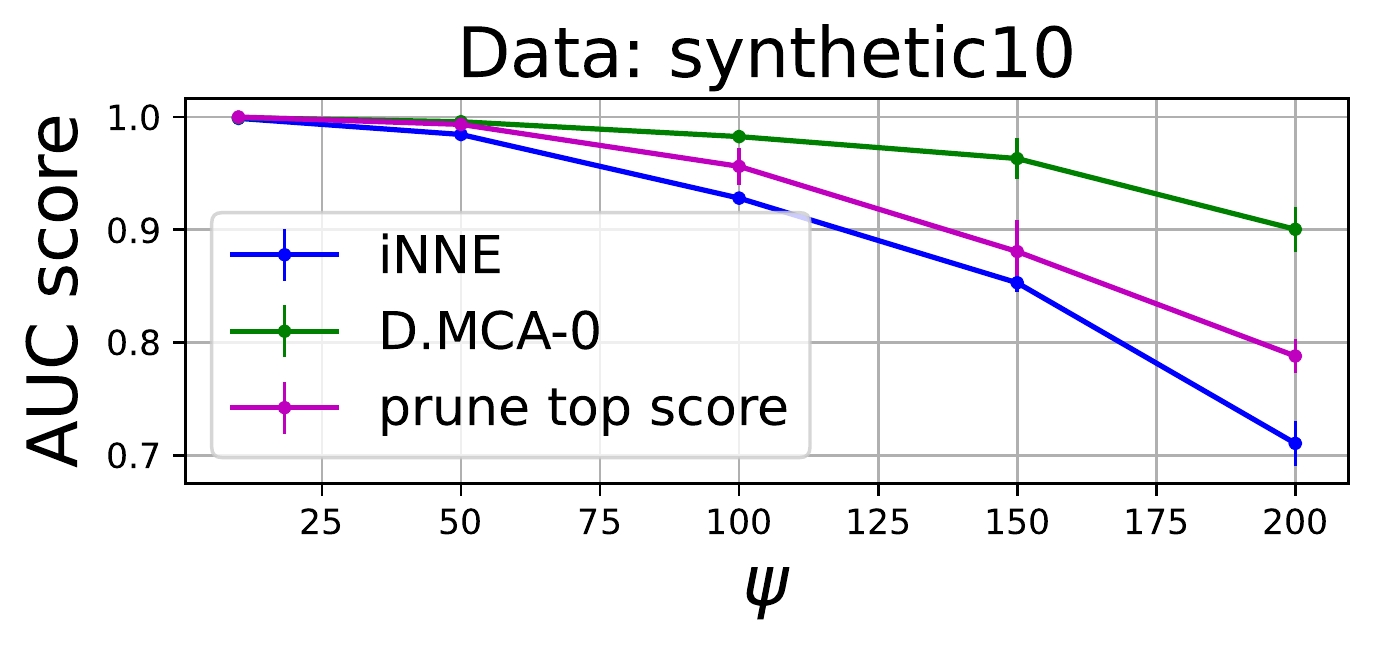}
    \end{subfigure}
    \hfill
    \begin{subfigure}[b]{0.475\linewidth}
        \centering 
        \includegraphics[width=\linewidth]{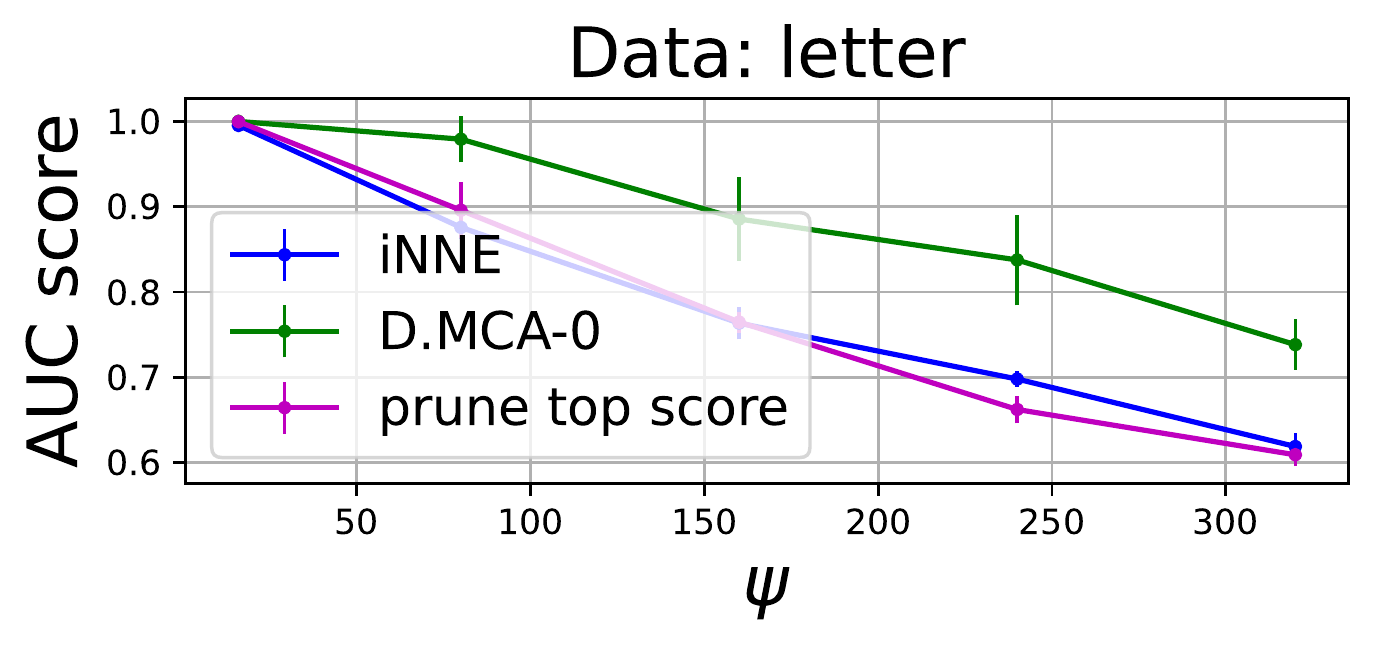}
    \end{subfigure}
    \caption{\textbf{Careful pruning of outliers is important}: simply pruning the top-$p$ scoring points (i.e., prune\_top\_score) may even decrease the accuracy of the base detector iNNE. Thanks to careful pruning, our \delac performs much better.}
    \label{fig:motivation_puning}
\end{figure}

Here we evaluate the ability of our methods to prune outliers out of the set~$\gR_{\psi}$ of subsampled points, and thus to avoid masking.
We begin by justifying the need for a carefully designed pruning procedure.
Fig. \ref{fig:motivation_puning} reports AUC score versus $\psi$ for datasets \texttt{synthetic10} and \texttt{letter}.
We consider three methods:
\begin{enumerate*}[label=(\alph*)]
\item \delaczero: with our proposed pruning procedure;
\item iNNE: the baseline that does not prune outliers, and;
\item prune\_top\_score: an alternative method created from \delaczero with the use of a substitute pruning procedure that simply removes from $\gR_{\psi}$ the top-$p$ scoring points of the previous iteration.
\end{enumerate*}
Note that \delaczero is used in this experiment to avoid any bias that could be caused by \delac's ``warm-up'' phase. 
As it can be seen, \delaczero clearly wins; the other two competitors alternate among the second and third positions.
We made a careful analysis to find out why method prune\_top\_score is not effective. 
We found that its simple pruning procedure avoids masking at the cost of overfitting the model with a strong tendency to have in $\gR_{\psi}$ only points from the core of the inlier clusters.
Similar results were obtained on our other datasets, which are omitted for brevity.

\begin{figure}[!ht]
    \begin{subfigure}[b]{0.475\linewidth}
        \centering 
        \includegraphics[width=\linewidth]{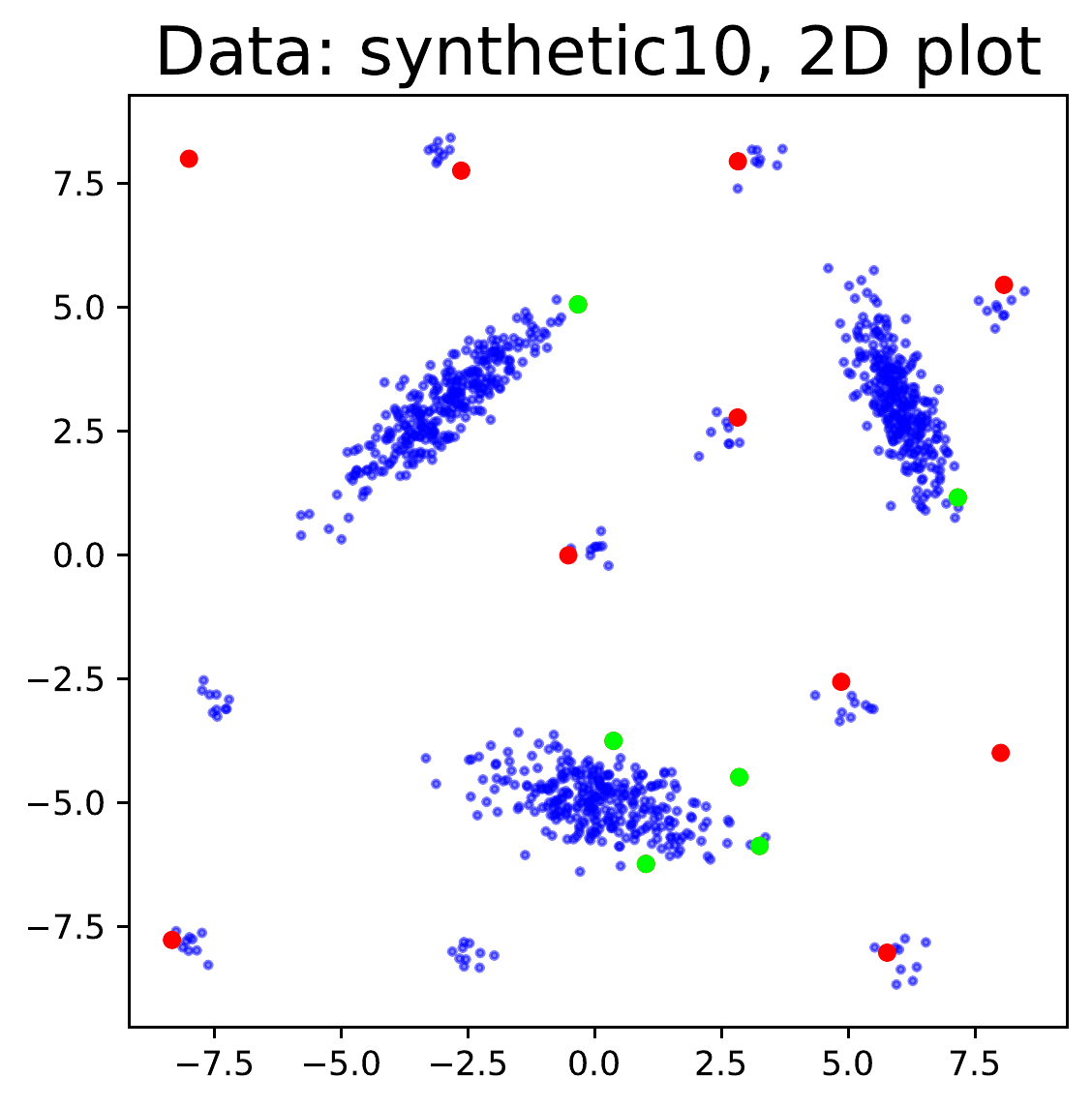}
    \end{subfigure}
    \hfill
    \begin{subfigure}[b]{0.475\linewidth}
        \centering 
        \includegraphics[width=\linewidth]{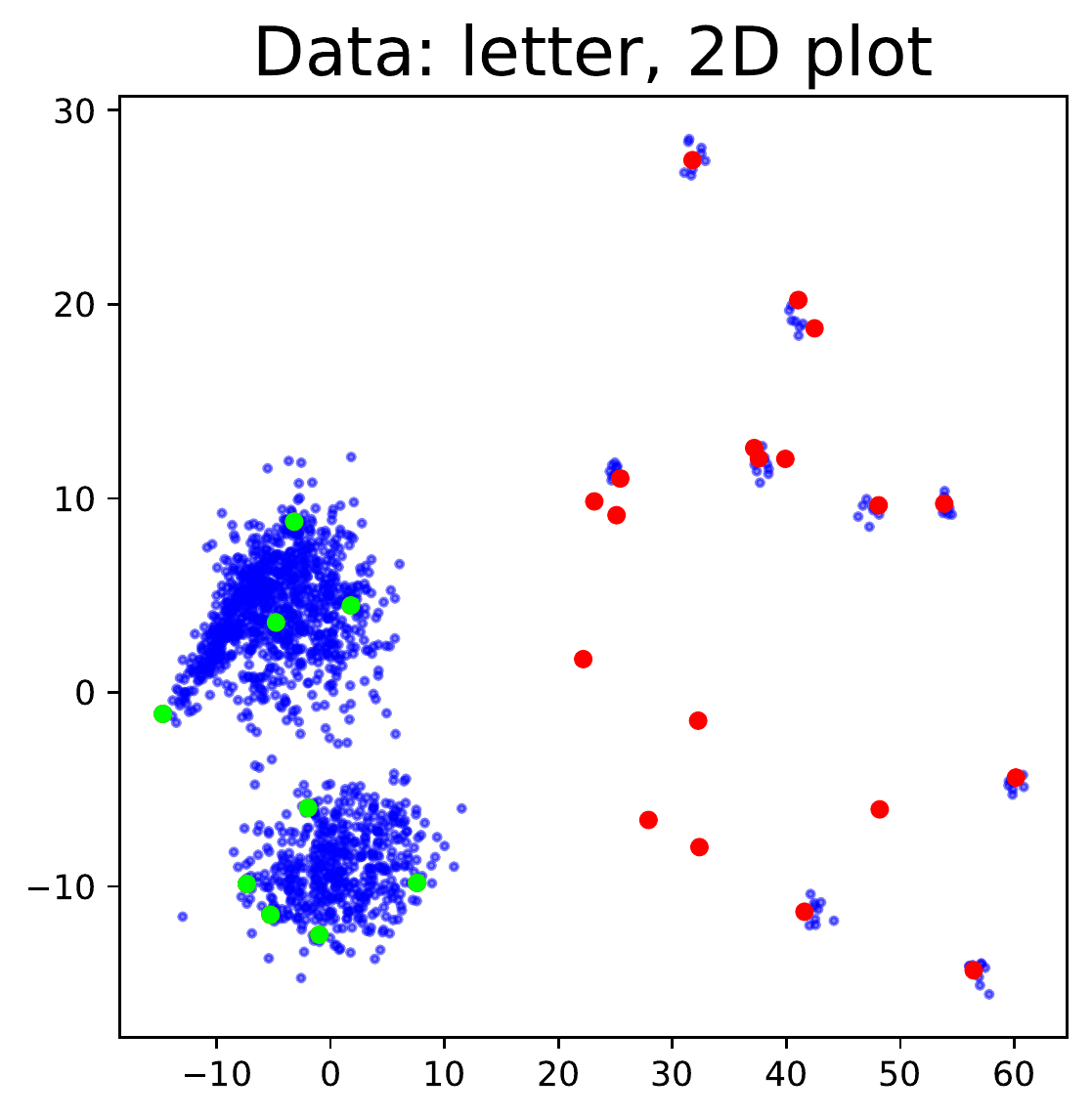}
    \end{subfigure}
    \caption{False Positive data points on the 2D plots of \texttt{synthetic10}, and \texttt{letter} (PCA projection) at iteration $10$ where $t=100, \psi = 16$. \textcolor{green}{Green} denotes false positive points, \textcolor{blue}{blue} denotes inlier points and \textcolor{red}{red} denotes true outliers.}
    \label{fig:clothes_lines_2d_data_synthetic10}
\end{figure}

\begin{figure}[]
    \centering
    \begin{subfigure}[b]{0.48\linewidth}
        \centering
        \includegraphics[width=\linewidth]{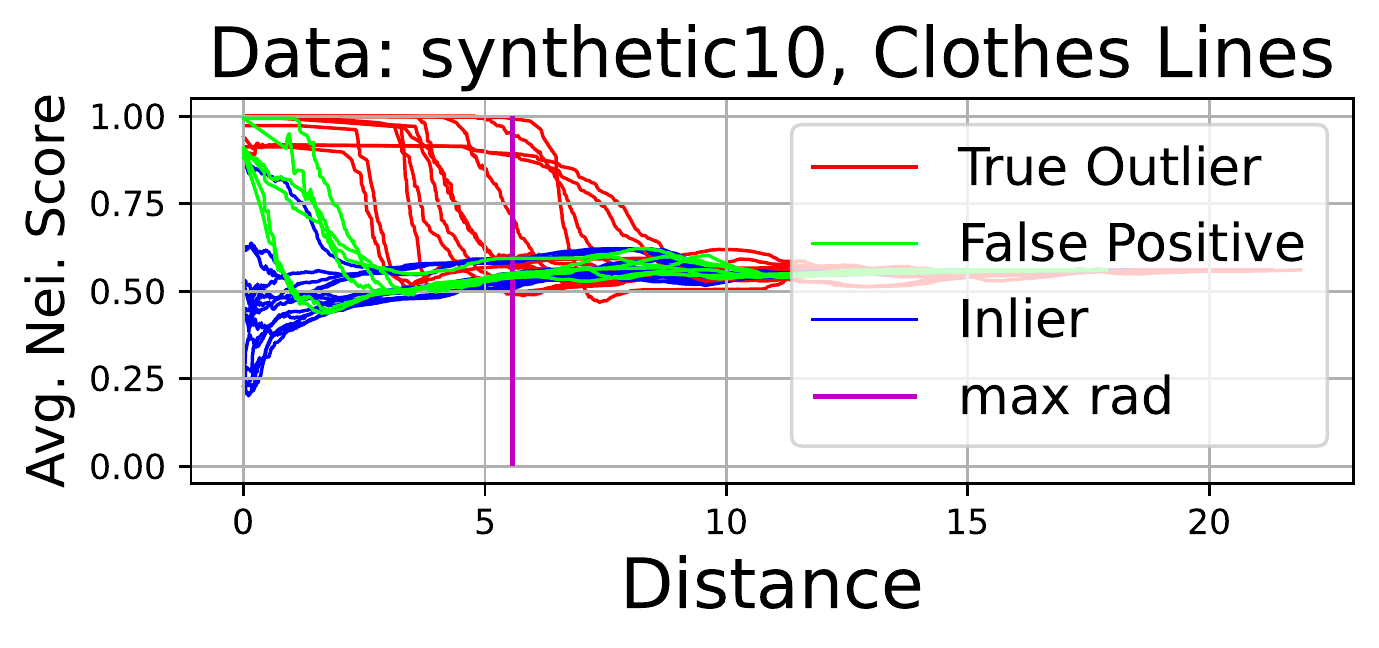}
    \end{subfigure}
    \hfill
    \begin{subfigure}[b]{0.48\linewidth} 
        \centering 
        \includegraphics[width=\linewidth]{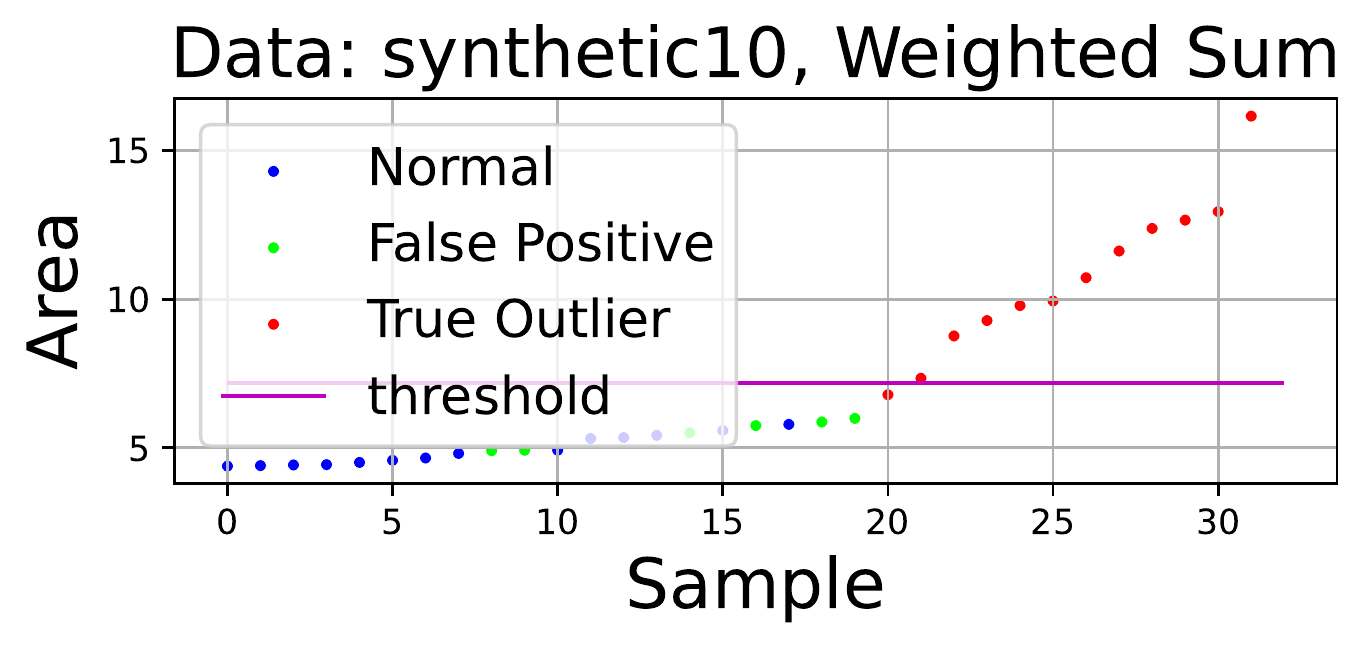}
    \end{subfigure}
   \vskip\baselineskip
    \begin{subfigure}[b]{0.48\linewidth}
        \centering 
        \includegraphics[width=\linewidth]{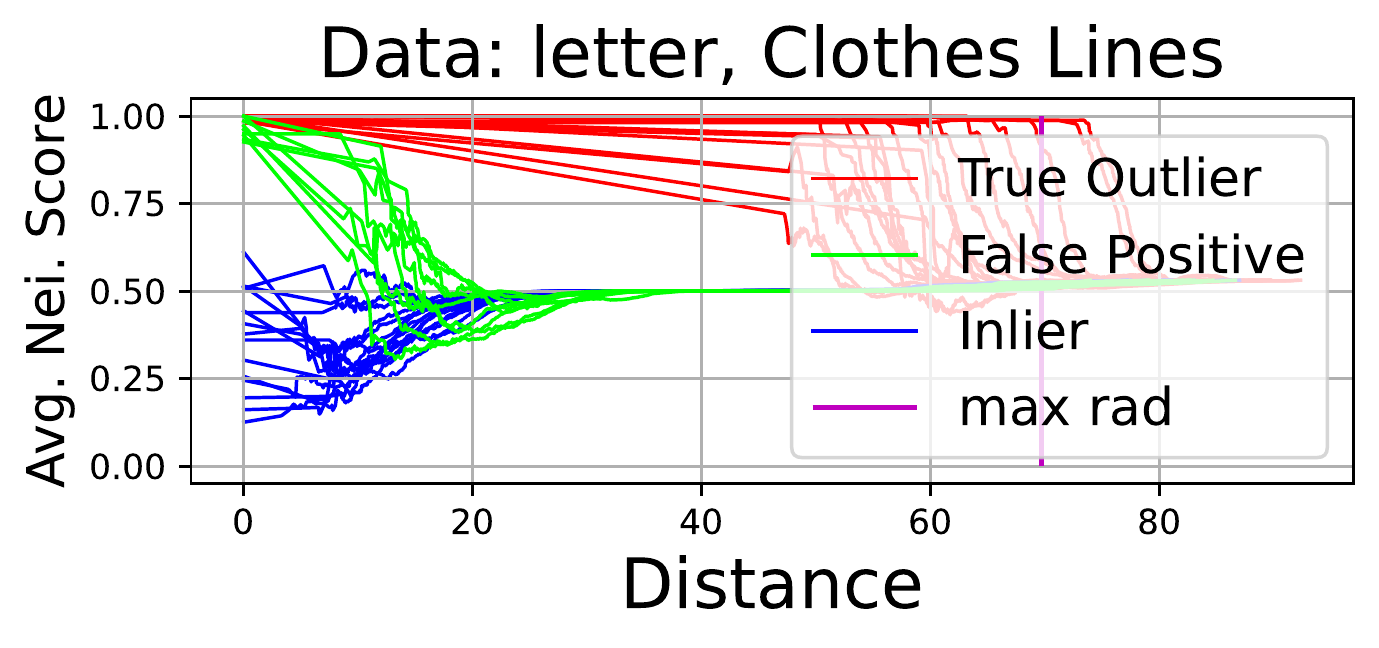}
    \end{subfigure}
    \hfill
    \begin{subfigure}[b]{0.48\linewidth}
        \centering 
        \includegraphics[width=\linewidth]{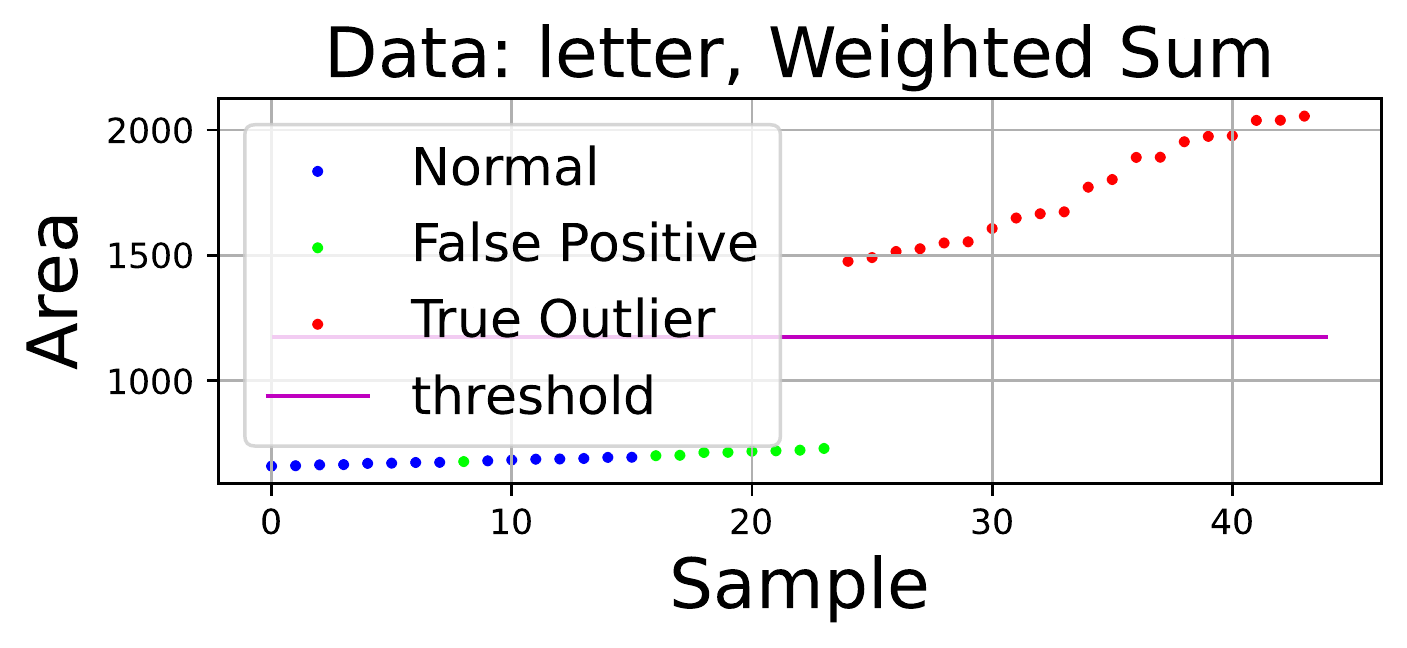}
    \end{subfigure}
    \caption{Clothes-lines and weighted area under clothes-lines on \texttt{synthetic10} at model/iteration 10 where $t=100, \psi = 16$. \textcolor{green}{Green} denotes false positive points, \textcolor{blue}{blue} denotes inliers and \textcolor{red}{red} denotes true outliers.}
    \label{fig:clothes_lines_FP_defense_synthetic10}
\end{figure}

These results lead us to conclude that carefully designed pruning of outliers is very important, as the use of a straightforward pruning procedure can actually degrade the scores of the baseline iNNE.
Then, we move on to detailing how our proposed pruning procedure behaves.
Fig.~\ref{fig:clothes_lines_2d_data_synthetic10} highlights the true outliers (in red) and the false positives (in green) in
the representative set $\gH$ for datasets~\texttt{synthetic10} and \texttt{letter}; the remaining points (not in $\gH$) are in blue.
As \texttt{letter} is $32$-dimensional, its PCA 2D projection is shown.
We consider the 10-th iteration, and hyperparameter values~$t=100$ and $\psi=16$.
As it is expected, $\gH$ has mostly true outliers, but a few false positives do exist.
Fig.~\ref{fig:clothes_lines_FP_defense_synthetic10} illustrates the corresponding clothes lines (left), and their weighted sums (right).
Note that the true outliers in fact have considerably larger weighted sums than the false positives.
Thanks to this distinguishing quantity, our carefully designed pruning procedure correctly identifies and filters out all false positives.

\begin{figure}[!t]
    \centering
    \begin{subfigure}[b]{0.48\linewidth}
        \centering
        \includegraphics[width=\linewidth]{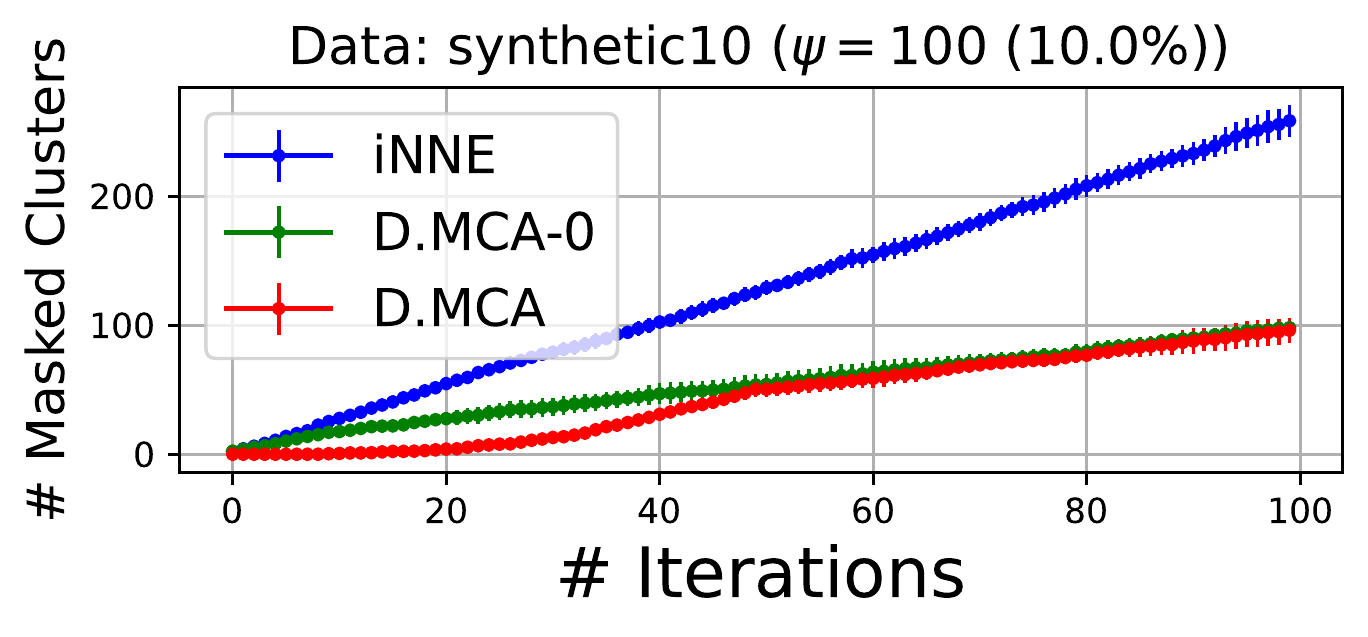}
    \end{subfigure}
    \hfill
    \begin{subfigure}[b]{0.48\linewidth}  
        \centering 
        \includegraphics[width=\linewidth]{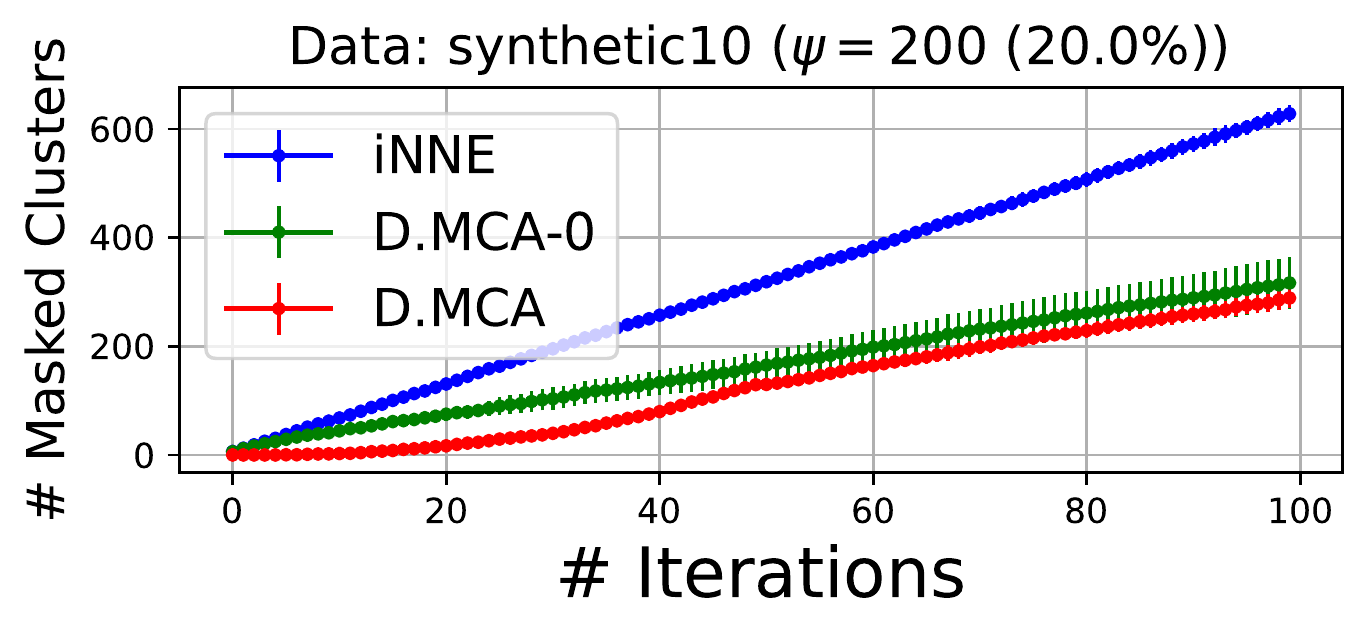}
    \end{subfigure}
   \vskip\baselineskip
    \begin{subfigure}[b]{0.475\linewidth}   
        \centering 
        \includegraphics[width=\linewidth]{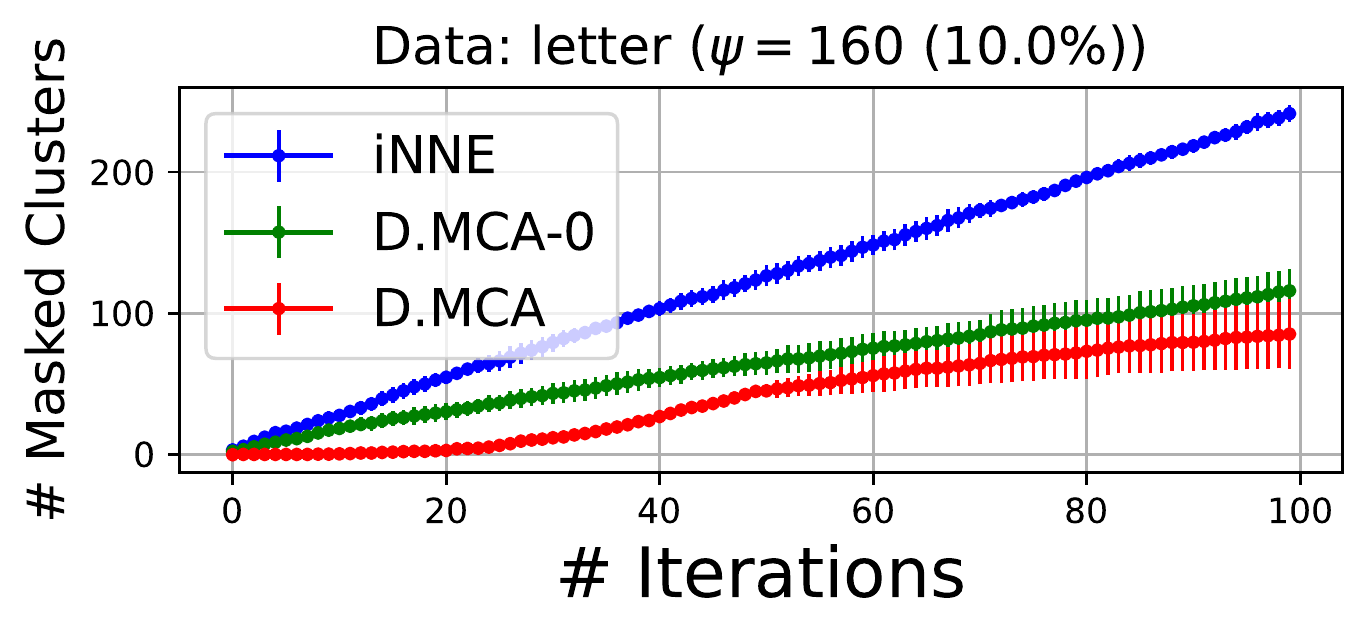}
    \end{subfigure}
    \hfill
    \begin{subfigure}[b]{0.475\linewidth}
        \centering 
        \includegraphics[width=\linewidth]{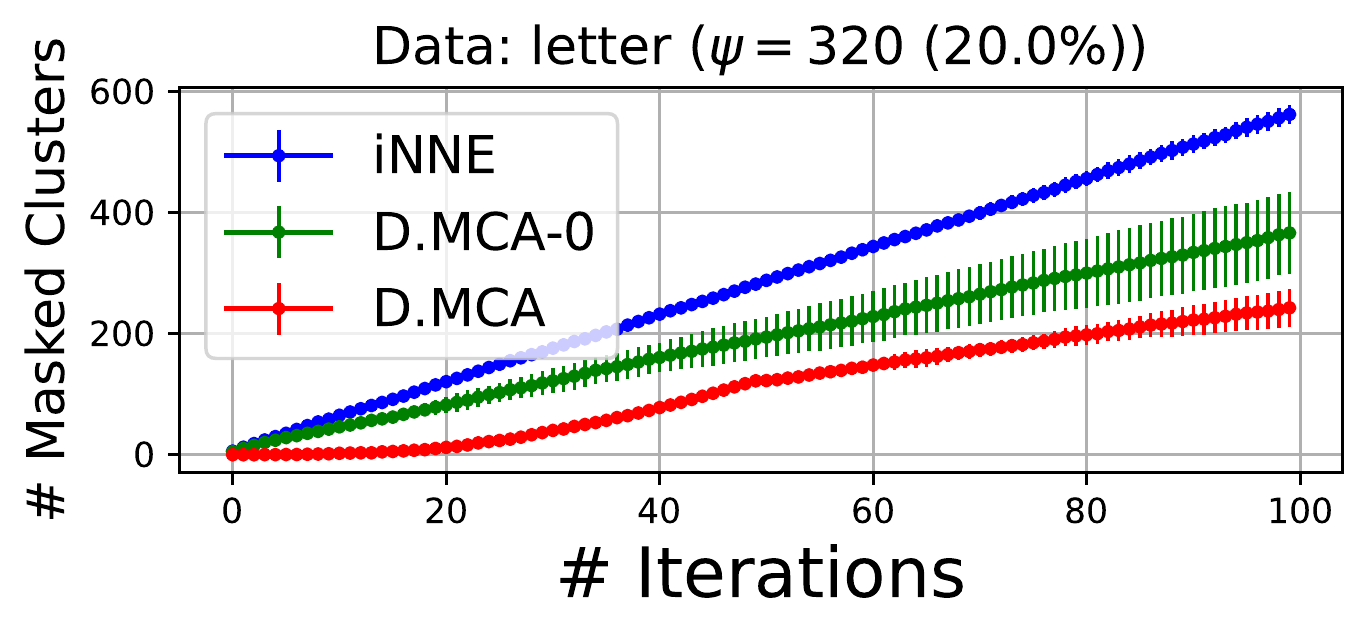}
    \end{subfigure}
    \caption{A comparison of cumulative occurrences of masking over 100 iterations/models for \textcolor{blue}{iNNE}, \textcolor{green}{\delaczero} and \textcolor{red}{\delac} on \texttt{synthetic10} and \texttt{letter} (the lower the better).}
    \label{fig:cumulative_masking}
\end{figure}

\begin{figure}
    \begin{subfigure}[b]{0.475\linewidth}
        \centering 
        \includegraphics[width=\linewidth]{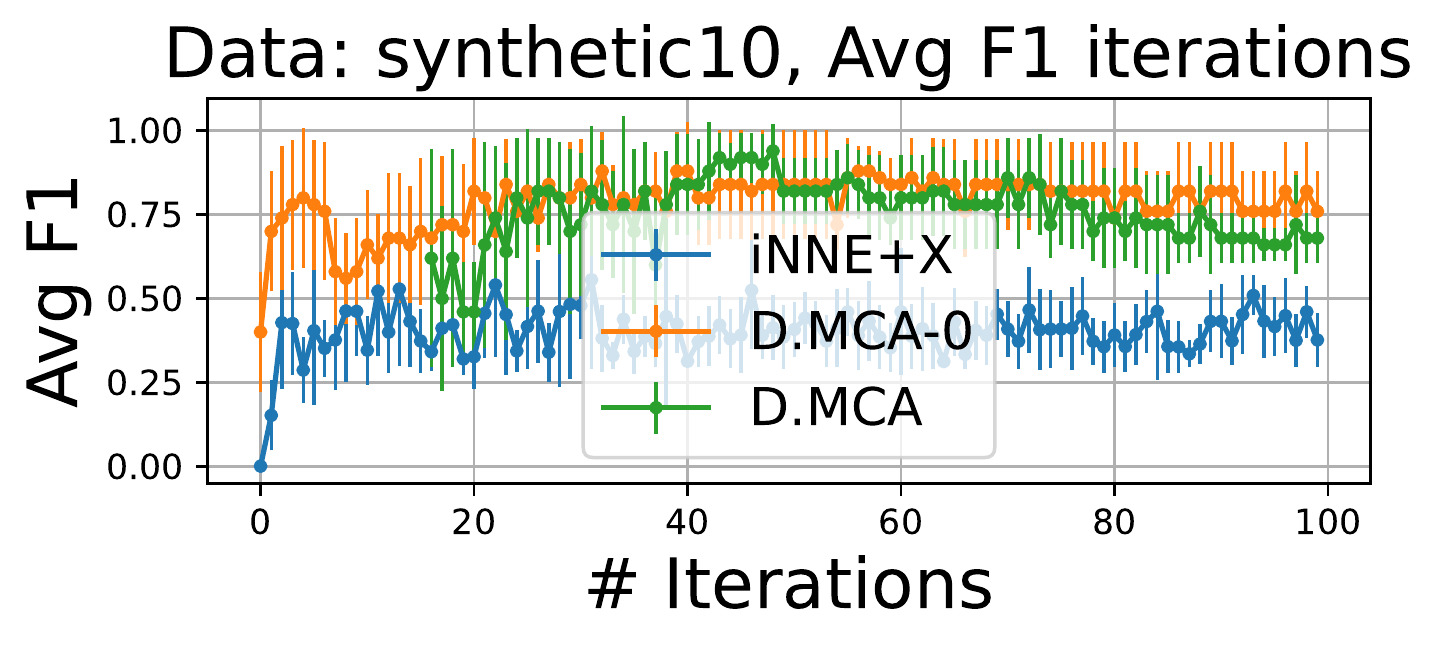}
    \end{subfigure}
    \hfill
    \begin{subfigure}[b]{0.475\linewidth}
        \centering 
        \includegraphics[width=\linewidth]{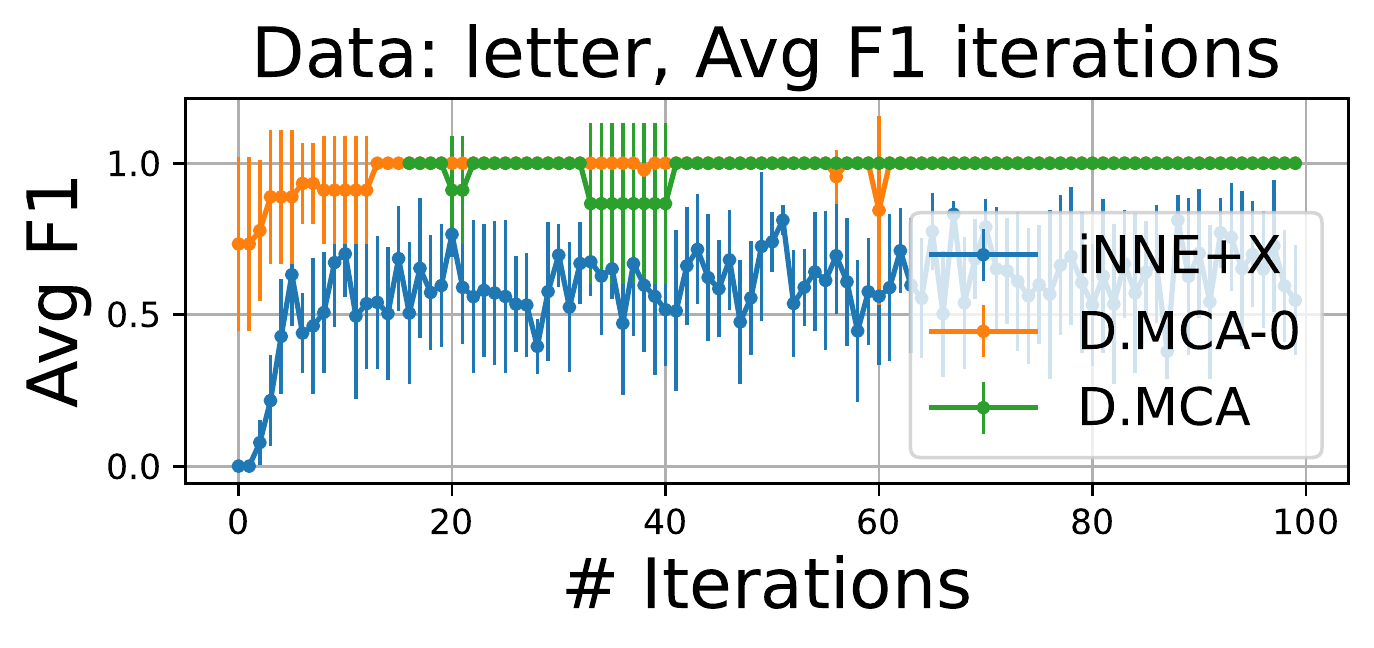}
    \end{subfigure}
    
    \vskip\baselineskip
    \begin{subfigure}[b]{0.475\linewidth}
        \centering 
        \includegraphics[width=\linewidth]{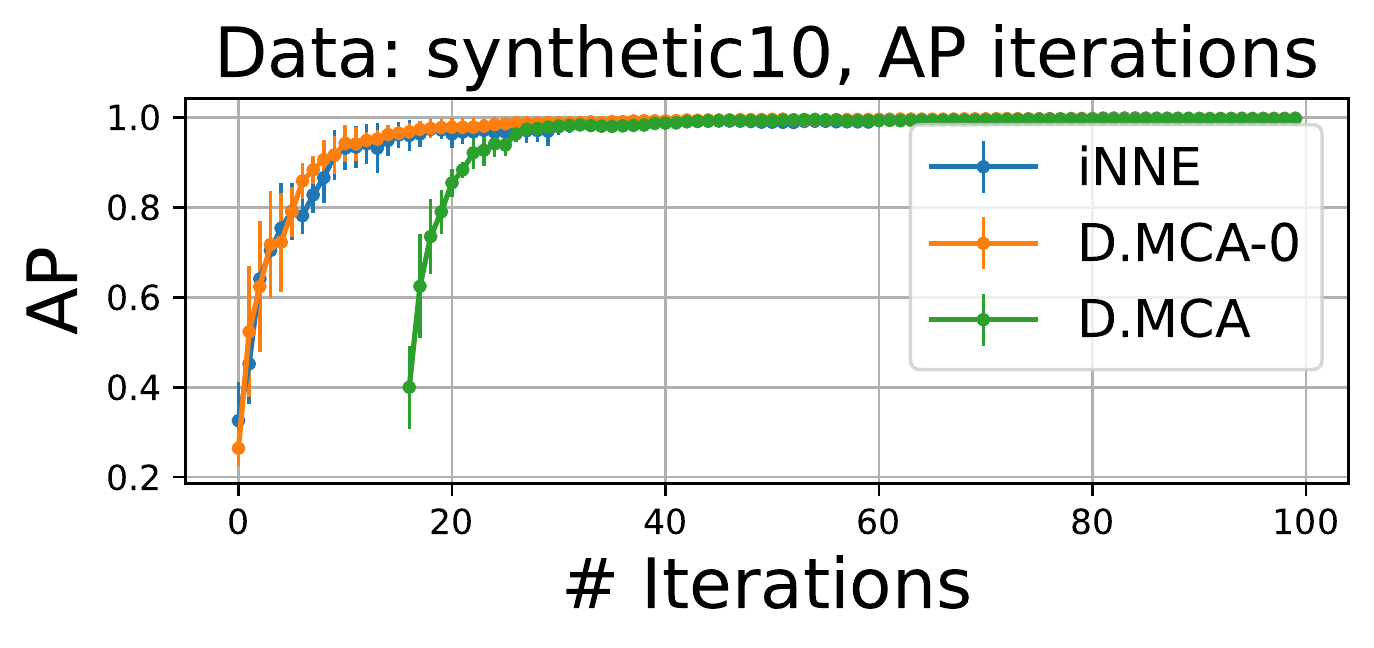}
    \end{subfigure}
    \hfill
    \begin{subfigure}[b]{0.475\linewidth}
        \centering 
        \includegraphics[width=\linewidth]{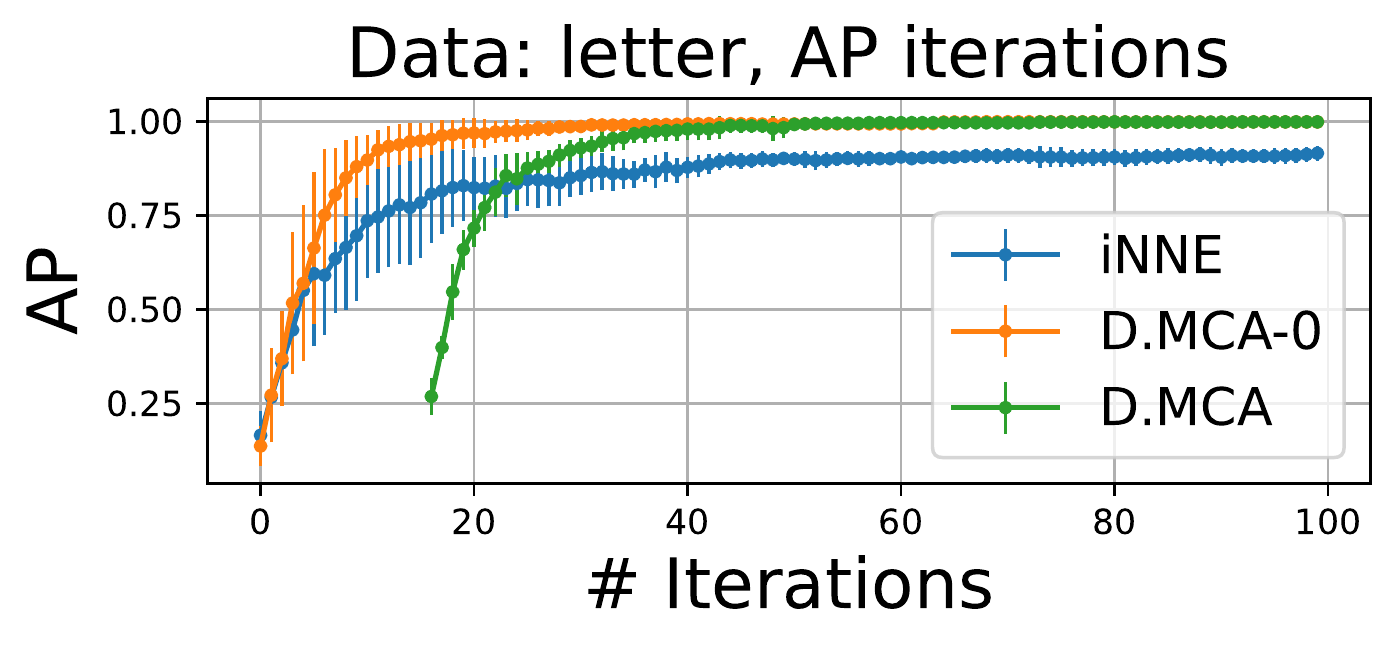}
    \end{subfigure}
    
   \vskip\baselineskip
    \begin{subfigure}[b]{0.475\linewidth}
        \centering 
        \includegraphics[width=\linewidth]{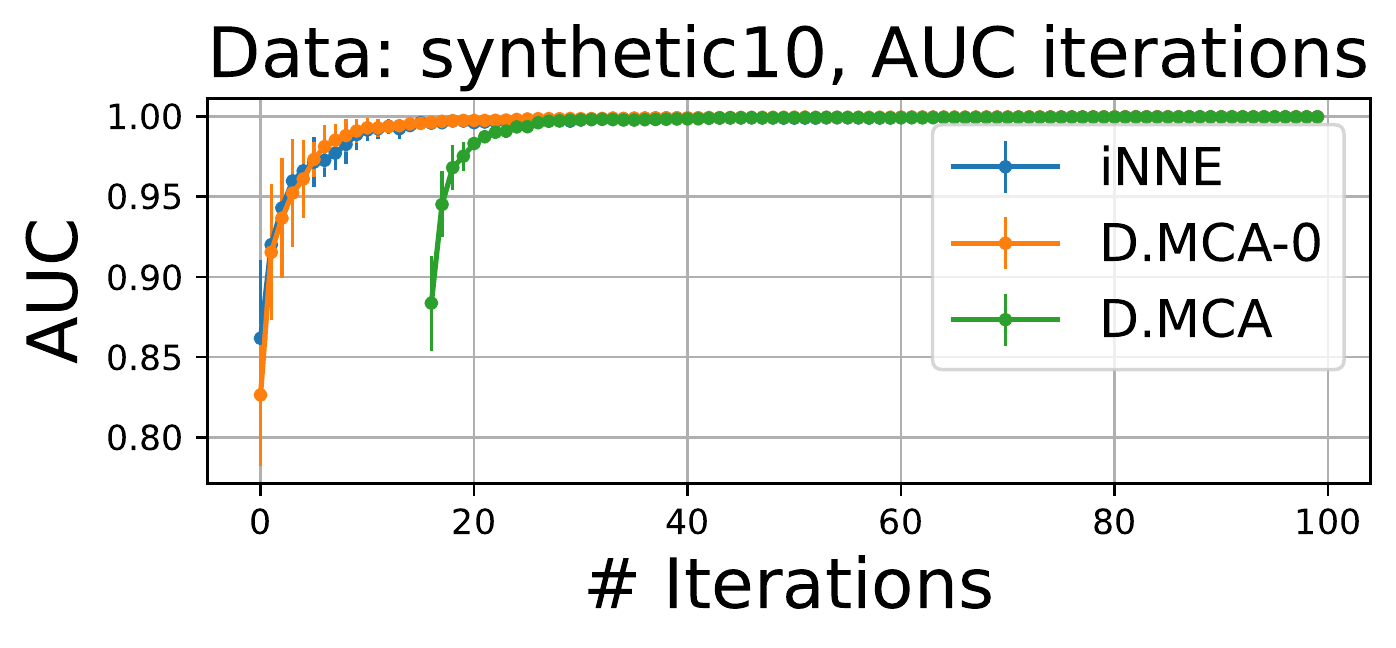}
    \end{subfigure}
    \hfill
    \begin{subfigure}[b]{0.475\linewidth}
        \centering 
        \includegraphics[width=\linewidth]{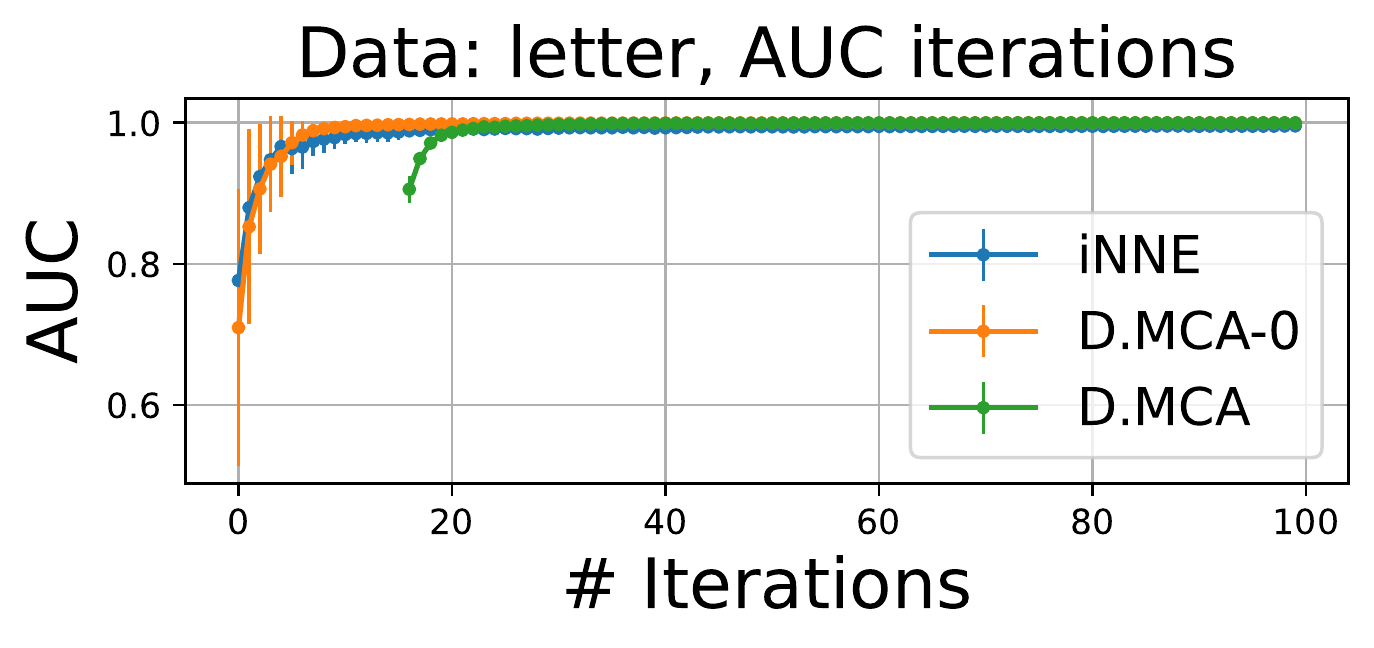}
    \end{subfigure}
    
    \caption{Performance comparison across iterations. 
    We plot the performance of \delac in Phase 2 only, i.e. starting at the 16-th iteration for $\psi_{\max} =16$, since \delac assigns outlier scores in Phase 2 only.
    }
    \label{fig:iteration_plots}
\end{figure}

Fig.~\ref{fig:cumulative_masking} further corroborates this fact by reporting the cumulative counts of masked clusters over iterations for \delac, and \delaczero. 
Datasets~\texttt{synthetic10}, and \texttt{letter} are used with varying $\psi$, and $t=100$.
We also report results for the baseline iNNE with increasing numbers of models.
As it can be seen, our methods are both considerably more successful than iNNE in avoiding masking; and, \delac performs even better than \delaczero in this task.
Finally, Fig.~\ref{fig:iteration_plots} confirms that mask avoiding indeed translates to improved accuracy.
We report the corresponding results of assignment (Avg F1), and detection (AP, and AUC) for \delac, \delaczero, and iNNE.
X-means is used on top of iNNE for assignment.
Note that the accuracy of all methods increase with more iterations/models, where both our \delac and \delaczero are clearly better than iNNE in detection as well as in assignment.

\section{Related Works}
\label{sec:related_works}
\textbf{Point outlier detection.~}
Outlier detection on $d$-dimensional point-cloud data has a long literature, with a large pool of detectors.
Those can be organized by the type of modeling approach that they take, such as distance-based, density-based, statistical-, cluster-, angle-, depth-based \citeapp{chandola2009anomaly,han2012outlier} and most recently deep learning based methods \citeapp{chalapathy2019deep,journals/corr/abs-2009-11732,pang2021learning}.
They can also be categorized by the setting that they consider, such as context-driven \citeapp{liang2016robust,conf/pkdd/MeghanathPA18,song2007conditional}, streaming settings \citeapp{conf/ijcai/TanTL11,gupta2013outlier,salehi2016fast,conf/kdd/ManzoorLA18}, or distributed settings \citeapp{yan2017distributed,corain2021dbscout,sparx22}.

Detection methods in regards to different type of outliers have drawn relatively less attention.
Most detection models assume outliers to be scattered points. While some body of work has distinguished global (or gross) outliers from local (or subtle) outliers \citeapp{breunig2000LOF,he2003discovering,kriegel2009loop,dang2013local,salehi2016fast,yan2017distributed}, both classes of work focus on individual isolate outliers.  Differently, our work aims to detect both scattered (i.e., isolate) as well as clustered outliers; and beyond detection, where a set or ranked list of outliers are reported, we organize the non-isolate outliers into their respective (micro-)clusters.

\textbf{Collective or group outlier detection.~}
Relative to the vast body of work on general-purpose outlier detection at large, methods with special focus on detecting clustered outliers in point-cloud data are limited \citeapp{he2003discovering,jiang2008clustering,liu2010SCiForest,lee2021gen2out}.
A similar notion, that is, collective anomalies have been explored for data different from point-clouds, including sequence data \citeapp{sun2006mining,maru2020collective}, graph data \citeapp{ye2015discovering,hooi2016fraudar,feng2021eaglemine}, and spatial data \citeapp{shekhar2001detecting}.

We remark that many general-purpose detectors are readily capable of detecting clustered outliers if present, even though the method did not originally target detecting them per se. For example, LOF \citeapp{breunig2000LOF} could flag all members of an outlier cluster provided that its hyper-parameter $k$ is larger than the size of the cluster. Similarly, the multi-granularity detection algorithm LOCI \citeapp{papadimitriou2003LOCI} can provide a large outlier score to outlier cluster members as the cluster would stand out at some radius (i.e. granularity). The key difference here is that those general-purpose methods do not distinguish between the type of outliers, in other words, they cannot explicitly annotate which of their detected outliers are clustered, let alone assign (clustered) outliers to their respective clusters.

The closest to our present work is 
\gensqout~\citeapp{lee2021gen2out}, 
a two-stage approach that proceeds detection with post-hoc clustering by DBSCAN~\citeapp{ester1996dbscan}. A major challenge with an approach that employs a clustering algorithm is that it inherits the challenges of clustering associated with varying size, density, and number of clusters.
In fact, it inherits two hyperparameters of DBSCAN, $\epsilon$ and minPts, which are nontrivial to set. \gensqout uses the default values suggested by the original authors, which may be subpar from task to task.
In addition, a two-stage approach with two independent steps misses on the opportunity to leverage a synergy between detection and micro-cluster assignment. 
Put differently, downstream clustering  simply works with the detector's output however subpar the detection results might be, without any feedback loop. Differently, our proposed \delac tackles both problems alternatingly, allowing and leveraging a synergy between the two.

\section{Conclusion}
\label{sec:conclusion}
In this paper we considered the two-pronged problem of detecting (scattered and clustered) outliers {\em as well as} explicitly assigning the clustered outliers to respective micro-clusters. To this end we proposed \delac, which addresses both problems simultaneously in a synergistic fashion, as opposed to a two-stage approach that handles micro-clusters \textit{post hoc} as an independent task. \delac is a sequential ensemble, 
which relies on a subsample of the input data that is polluted with outliers. Its main idea is to iteratively prune the detected outliers to improve the representativeness of the subsample.
The synergy between the tasks occurs through the identification of outliers to prune prior to subsampling; where pruning outliers reduces the risk of ``masking'' leading to better identification of the outlier micro-clusters, which in turn enables better pruning.  
Importantly, we also designed \delac to be robust to the subsample size, a critical hyperparameter to balance representativeness and masking, where we employed a hyperensemble in a ``warm up'' phase to obtain a pruned dataset to start with.
Extensive experiments on 16 datasets against 8 SOTA detectors showed that \delac provides competitive detection performance. Of all detectors, only \gensqout is designed to also tackle micro-cluster assignment, however we found it to perform quite poorly on this task. While post hoc clustering remains a tricky task itself, we also showed that effective detectors paired with hyperparameter-free SOTA clustering algorithms produce subpar results. Overall, our proposed \delac leads as the SOTA approach to our two-pronged problem.

\bibliographystyleapp{IEEEtran}
\bibliographyapp{references}

\end{document}